\documentclass[]{antgroup}
\PassOptionsToPackage{numbers, compress}{natbib}
\usepackage{antgroup}

\usepackage{graphicx}
\usepackage{tikz}
\usepackage{todonotes}
\usepackage{multirow}
\usepackage{amsmath}
\usepackage{cleveref}
\usepackage{subcaption}
\usepackage{multicol}
\usepackage{amssymb}
\usepackage{array}
\usepackage{bm}
\usepackage{enumitem}
\usepackage{algorithm}
\usepackage{algpseudocode}
\usepackage{tabularx}
\usepackage{booktabs}
\usepackage{bbm}
\usepackage{makecell}
\usepackage{wrapfig}
\usepackage{colortbl}
\usepackage{threeparttable}
\usepackage{fontawesome5}
\usepackage[normalem]{ulem}
\usepackage{enumitem}
\usepackage{caption}
\usepackage{arydshln}
\usepackage{pifont}
\usepackage[T1]{fontenc}
\usepackage{listings}
\usepackage[most]{tcolorbox}
\usepackage{textcomp}
\usepackage{upquote}
\useunder{\uline}{\ul}{}

\usepackage{amsmath,amsfonts,bm}

\def\eqref#1{equation~\ref{#1}}
\def\1{\bm{1}}

\DeclareMathAlphabet{\mathsfit}{\encodingdefault}{\sfdefault}{m}{sl}
\SetMathAlphabet{\mathsfit}{bold}{\encodingdefault}{\sfdefault}{bx}{n}

\newcommand*\justify{%
  \fontdimen2\font=0.4em% interword space
  \fontdimen3\font=0.2em% interword stretch
  \fontdimen4\font=0.1em% interword shrink
  \fontdimen7\font=0.1em% extra space
  \hyphenchar\font=`\-% allowing hyphenation
}

\renewcommand{\texttt}[1]{%
  \begingroup
  \ttfamily
  \begingroup\lccode`~=`/\lowercase{\endgroup\def~}{/\discretionary{}{}{}}%
  \begingroup\lccode`~=`[\lowercase{\endgroup\def~}{[\discretionary{}{}{}}%
  \begingroup\lccode`~=`.\lowercase{\endgroup\def~}{.\discretionary{}{}{}}%
  \catcode`/=\active\catcode`[=\active\catcode`.=\active
  \justify\scantokens{#1\noexpand}%
  \endgroup
}

\usepackage{amsmath}
\usepackage{amssymb}
\usepackage{amsfonts}                               % blackboard math symbols
\usepackage{amsthm}
\usepackage[mathcal]{eucal}
\usepackage{mathrsfs}
\usepackage{bm}                                     % bm command
\usepackage{blkarray}                               % to support matrix
\usepackage{nicefrac}                               % compact symbols for 1/2, etc.

\usepackage{wrapfig}
\usepackage{graphicx}                               % include pdf figures
\usepackage{caption}
\usepackage{cleveref}
\usepackage{tikz}                                          
\usepackage{circuitikz}
\usetikzlibrary{patterns,snakes}
\usetikzlibrary{positioning,calc,fit,decorations.pathmorphing,shapes.geometric, shapes.gates.logic.US, calc}
\usetikzlibrary{arrows,arrows.meta,decorations.markings,shapes,shapes.arrows}
\usetikzlibrary{decorations,decorations.pathreplacing}
\usetikzlibrary{backgrounds}
\usepackage{filecontents}                           % support to pgfplots
\usepackage{pgfplots}
\usepackage{pgfplotstable}
\usepgfplotslibrary{groupplots}
\usepackage{scalefnt}
\pgfplotsset{compat=newest}
\usepackage{xcolor}

\definecolor{LightGray}{gray}{0.9}

\definecolor{myPink}{HTML}{EE6B98}
\definecolor{myBlue}{HTML}{6BB8FA}
\definecolor{myPurple}{HTML}{9F63F0}
\definecolor{myBlueBase}{HTML}{147BFF}

\definecolor{win}{HTML}{2E8B57}
\newcommand{\up}[1]{\textcolor{win}{\scriptsize (+#1)}}

\definecolor{nativegray}{HTML}{888888}
\definecolor{memblue}{HTML}{1F77B4}

\usepackage{xcolor}
\usepackage{xspace}

\definecolor{MorandiRed}{HTML}{B75E54}

\newcommand{\colorours}{\textsc{%
  \textcolor{MorandiRed}{M}%
  \textcolor{MorandiRed!90!black}{e}%
  \textcolor{MorandiRed!80!black}{m}%
  \textcolor{MorandiRed!70!black}{D}%
  \textcolor{MorandiRed!50!black}{r}%
  \textcolor{MorandiRed!40!black}{e}%
  \textcolor{MorandiRed!30!black}{a}%
  \textcolor{MorandiRed!20!black}{m}%
  \textcolor{MorandiRed!10!black}{e}%
  \textcolor{MorandiRed!0!black}{r}%
}\xspace}

\newcommand{\ours}{\textsc{MemDreamer}\xspace}

\title{\colorours: Decoupling Perception and Reasoning \\ for Long Video Understanding via Hierarchical Graph Memory and Agentic Retrieval Mechanism\thanks{~~Equal Contribution. $^\dag$ Corresponding Author.}}
\vspace{-2ex}
\author{\centering
Cong Chen$^{1,2,*}$,
Guo Gan$^{2,*}$,
Kaixiang Ji$^{1,*}$,
ZhaoYang Zhang$^1,^3$,
Zhen Yang$^4$,
\\
Guangming Yao$^1$,
Hao Chen$^2$,
Jingdong Chen$^1$,
Yi Yuan$^{1}$,
Chunhua Shen$^{5,2,1\dag}$,
}
\affiliation{$^1$Ant Group, ~~~ $^2$Zhejiang University, ~~~ $^3$Central South University \\$^4$HKUST (GZ), ~~~ $^5$Zhejiang University of Technology
}

\begin{document}
\maketitle

\vspace{-3ex}
\begin{abstract}
Current Vision-Language Models struggle with hours-long videos because processing full-length visual sequences induces prohibitive token explosion and attention dilution. To overcome this, we introduce \colorours to decouple perception and reasoning, shifting long-video understanding into an agentic exploration process. As a plug-and-play framework, it incrementally streams videos to construct a Hierarchical Graph Memory, a top-down three-tier architecture for semantic abstraction, anchored by a foundational graph capturing spatiotemporal and causal relations. During inference, the reasoning model employs agentic tool-augmented retrieval, navigating hierarchies, searching nodes, and traversing logical edges via an Observation-Reason-Action loop. Experiments show \colorours achieves SOTA results across four mainstream benchmarks, narrowing the gap with human experts to only 3.7 points. It constrains the reasoning context window to merely 2\% of full-context ingestion while delivering a 12.5 point absolute accuracy gain. Furthermore, statistical analysis uncovers a strong positive linear correlation between an VLM's performance on logic reasoning and long-video understanding benchmarks, establishing agentic capability scaling as a new paradigm for multimodal comprehension.

\textbf{Project:} \url{https://aim-uofa.github.io/MemDreamer/}

\end{abstract}

\vspace{-3ex}

\begin{figure}[hb]
    \centering
    \includegraphics[width=0.43\linewidth]{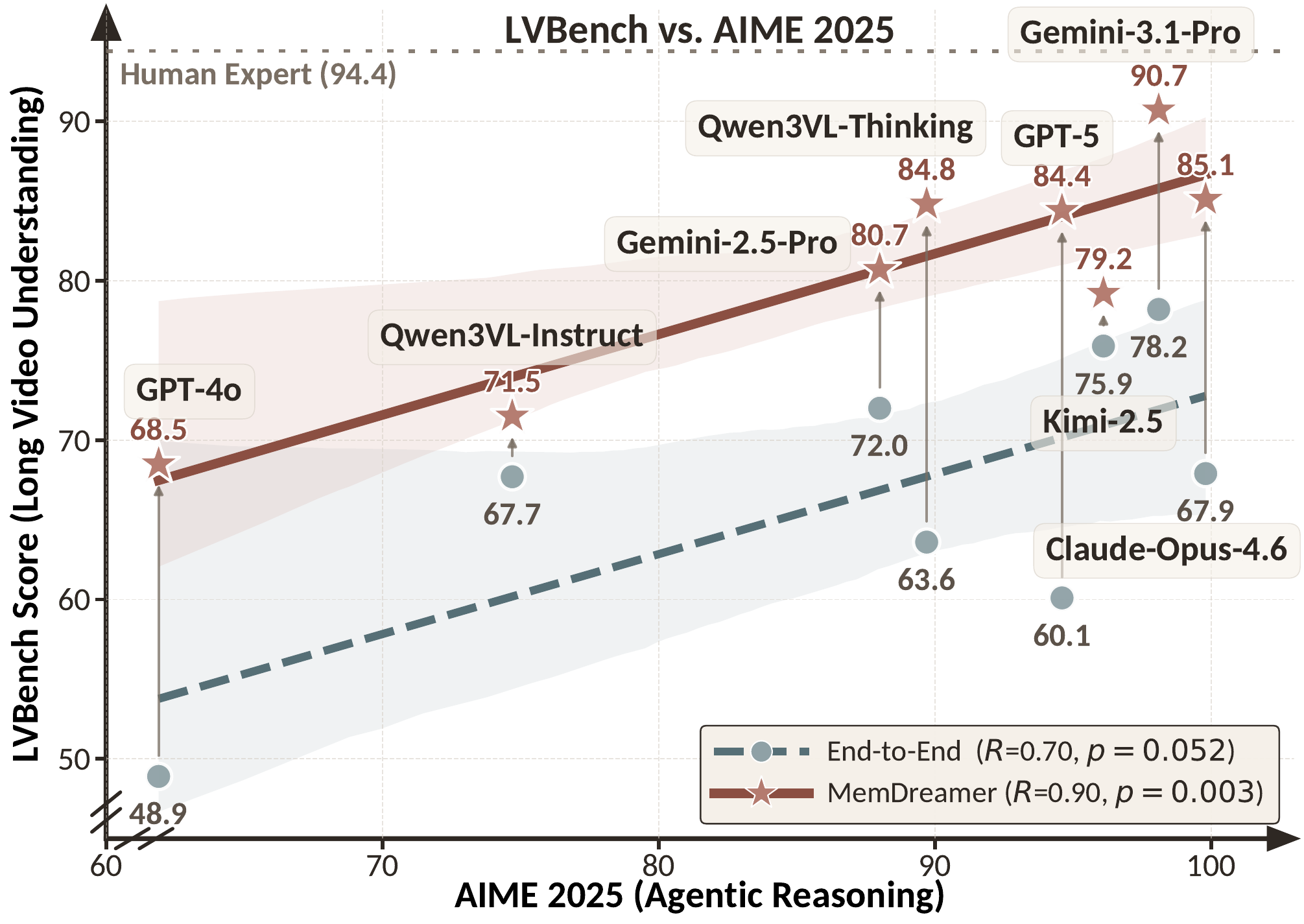}
    \includegraphics[width=0.35\linewidth]{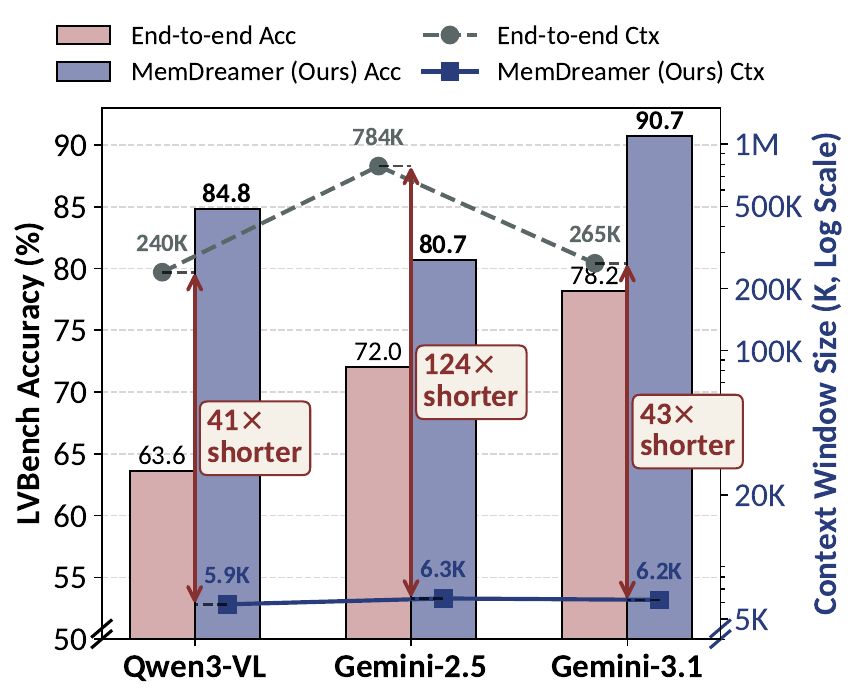}
    \caption{\footnotesize
        \textbf{(left) Decoupling perception unlocks reasoning ability for long video understanding.} We evaluate the same backbone in two regimes: as an end-to-end VLM (\textcolor{gray}{gray}) and as the reasoning model of \colorours over a pre-built hierarchical graph memory (\textcolor{red}{red}). End-to-end performance is insensitive to reasoning ability, as the model is overwhelmed by noisy long-context tokens. In contrast, \colorours exhibits a strong linear trend, confirming that decoupling facilitates the direct transfer of the backbone's agentic reasoning ability into long video understanding performance.
        \textbf{(right) \colorours only needs a 5-6K reasoning window, 41-124$\times$ smaller than end-to-end input.}
    }
    \label{fig:teaser_combined}
\end{figure}

%\vspace{-3ex}

% \newpage
\section{Introduction}
Long video understanding~\citep{jin2024chatunivi,song2024moviechat} represents a core capability for driving Vision-Language Models (VLMs)~\citep{chen2024Internvl2.5,deepmind_gemini,gpt-4o} toward embodied intelligence and open-world interaction. Despite breakthroughs in single ~\citep{liu2023llava,chen2024internvl,chen2025perturbollava} or multi-images~\citep{wang2025uitars2,ye2025mobileagent,chen2025gui_shepherd,gan2026android_coach} and short video analysis~\citep{zhao2025mmvu}, current VLMs~\citep{li2024llamavid} struggle to scale to hours-long videos. This failure stems from the structural intricacy and extreme temporal redundancy inherent in long videos, which submerge critical reasoning signals beneath massive noise.

Current VLMs~\citep{lin2024videollava,zhang2024llavavideo,li2025videochat} typically adopt a tightly coupled strategy, relying on ultra-long context windows to process visual perception and logical reasoning simultaneously. This coupled paradigm flattens long videos into massive token streams via brute-force frame sampling, introducing two bottlenecks. \textbf{In perception}, it incurs an intractable token explosion: sampling a 2-hour video at 1 FPS generates over 1.6M tokens, drastically exceeding current context limits. \textbf{In Reasoning}, the influx of redundant tokens induces severe attention dilution~\citep{xiao2024sink} and exacerbates the ``lost in the middle'' phenomenon~\citep{liu2024lost}, significantly degrading long-range reasoning capabilities.

To overcome these bottlenecks, we introduce \colorours, a novel paradigm that decouples perception from reasoning. Specifically, our framework employs a perception model~\citep{deepmind_gemini,bai2025qwen3vl} to process videos in a streaming fashion, incrementally constructing a persistent memory bank. During inference, a separate reasoning model executes a search loop within this memory to retrieve task-relevant cues. While this streaming-and-retrieval mechanism circumvents context limits and token redundancy, its efficacy depends on the memory's structural organization. Existing flat~\citep{yang2025vca} or chunk-based~\citep{zhang2026dvd,wang2025videotree} storage schemes obscure global perspectives and sever temporal-causal links, causing decoupled reasoning to degenerate into myopic, exhaustive retrieval.

\begin{figure}[t]
  \includegraphics[width=\linewidth]{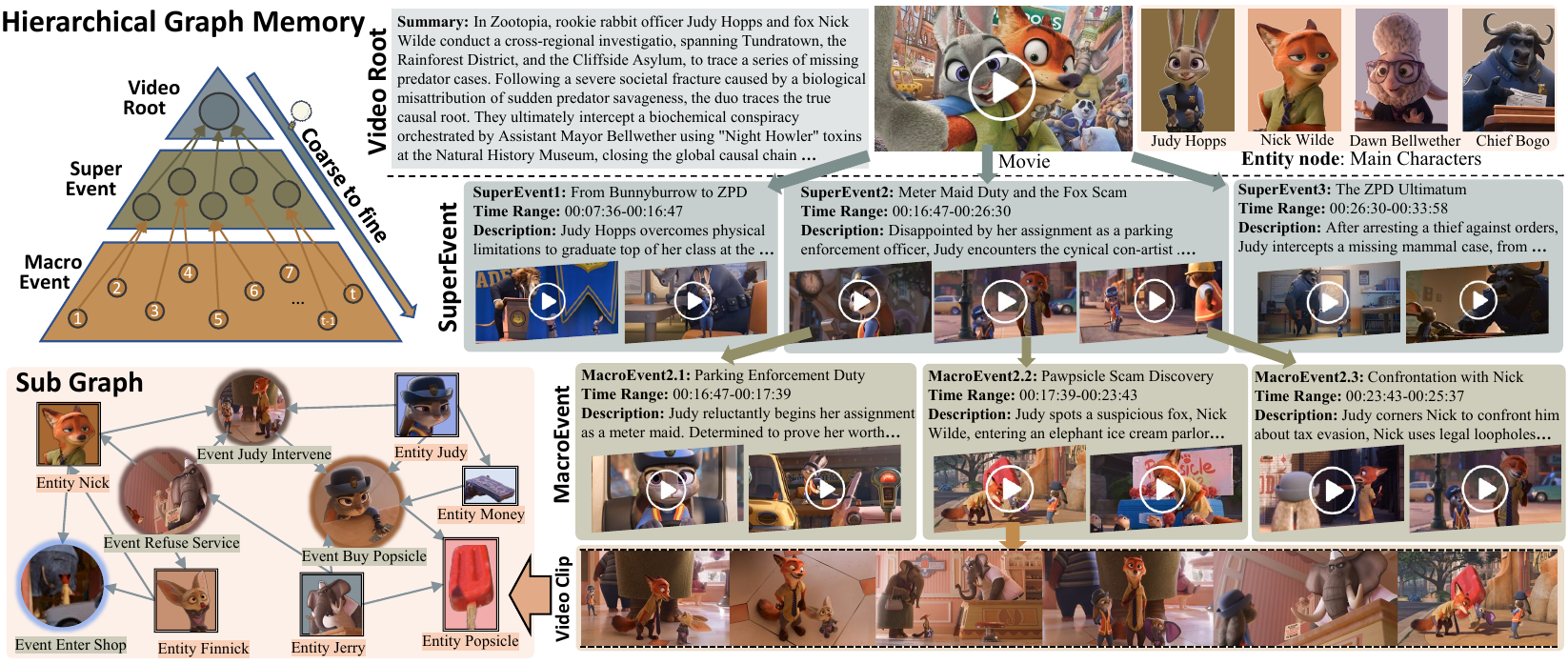}
  \caption{
    \colorours organizes a long video as a three-tier coarse-to-fine hierarchy with cross-tier topological edges; at the leaf tier, each segment is a subgraph of entities and micro-events via attributive and causal edges.
  }
  \label{fig:teaser}
\end{figure}

To bridge this gap, \colorours introduces a novel Hierarchical Graph Memory. 
Video content is inherently hierarchical~\citep{chen2025hieratok}, mirroring how humans comprehend a long video coarse-to-fine, starting from the overall plot, then scene by scene, event by event, finally down to individual actions, rather than memorizing a flat, exhaustive sequence of moments~\citep{pang2025mr.video,long2025m3agent}.
Moreover, these fine-grained events and entities are far from independent: they are intertwined by spatiotemporal and causal relations~\citep{edge2024graphrag} that no purely sequential representation can faithfully capture.
To realize this hierarchical and topological intuition, we construct a three-tier memory architecture shown in Figure~\ref{fig:teaser}. The top tier is a Video Root that summarizes global context, which is progressively decomposed into Super Events at the intermediate layer and Macro Events at the bottom. At the Macro Event tier, information is instantiated into a local subgraph characterizing entities, events, and their logical relations. This representation suppresses irrelevant detail at the appropriate granularity while preserving the long-range dependencies that downstream reasoning relies on.

To fully exploit this architecture, we propose a tool-augmented Agentic Retrieval Mechanism, superseding traditional full-context ingestion or similarity-based retrieval~\citep{lewis2020rag1,gao2023rag2}. We construct a dedicated Agentic Tool Bank comprising three categories: Navigation for vertical hierarchical traversal, Search for rapid node localization, and Graph Traversal for tracking logical chains along topological edges. By engaging in an iterative Observation-Reason-Action loop~\citep{yao2022react}, the model dynamically interacts with the memory. This design shifts long video understanding from passive token consumption to a multi-step active agentic exploration task.

Extensive experiments validate the superiority of \colorours. On challenging hours-level benchmarks~\citep{lvbench,longvideobench,videomme}, \colorours consistently achieves SOTA results~\citep{egoschema}, narrowing the performance gap with human experts to 3.7 points. 
Notably, our decoupled strategy achieves an absolute gain of 12.5 points over end-to-end coupled paradigms using only 2\% of the context window on hours-long video bench LVBench~\citep{lvbench}.
Furthermore, \colorours reveals a critical observation: we empirically establish a strong positive correlation between an VLM's intrinsic agentic capability and its long-video understanding performance, suggesting the scaling of agentic capacity as a new paradigm for future research.

In summary, our contributions are threefold:

1. \colorours decouples perception and reasoning via persistent memory and agentic retrieval, bypassing long-context bottlenecks. To our knowledge, we reveal for the first time a positive correlation between long-video performance and reasoning, suggesting a new optimization direction.

2. \colorours achieves SOTA performance across four representative long-video benchmarks. Using identical base models, we achieve substantial gains over traditional end-to-end baselines, narrowing the gap toward human expert levels.

3. \colorours establishes foundational design principles for multimodal memory systems. Empirically, hierarchical graphs and agentic tool-use outperform flat chunk-based storage and naive full-context ingestion or semantic matching, providing a cornerstone for future architectures.

\section{Related Work}
\subsection{Long Video Understanding}
Long video understanding remains a frontier for VLMs, marking the evolutionary leap to complex, real-world visual streams. To handle potentially infinite inputs, previous research has attempted to optimize visual encoders~\citep{lin2024videollava,zhang2024llavavideo} and scale native context windows~\citep{bai2025qwen3vl,gpt-4o,team2023gemini2.0flash}. However, even with leading models expanding context limits to 1M tokens like Gemini-2.5-Pro~\citep{gemini2.5}, systems still rely on aggressive frame subsampling to accommodate lengthy videos. This brute-force paradigm leaves models vulnerable to ``needle-in-a-haystack'' retrieval failures~\citep{wang2024needlemm,wang2026mmlongbench} and the ``lost in the middle'' phenomenon~\citep{liu2024lost} on benchmarks like LVBench~\citep{lvbench}. 
To bypass these bottlenecks, a more scalable paradigm structures video information within an external memory bank, transforming static perception into an agentic reasoning process.

\subsection{Agentic Memory Systems}
While memory systems like MemGPT~\citep{packer2023memgpt} and MemoryBank~\citep{zhong2024memorybank} achieve remarkable success in LLMs, adapting them for VLMs remains challenging. Early video attempts~\citep{song2024moviechat} rely on simple frame buffers, and subsequently M3-Agent~\citep{long2025m3agent} and WorldMM~\citep{yeo2025worldmm} encapsulate semantics into discrete bins. Yet, these flat designs lack top-down taxonomy and topological edges. Recently, VideoARM~\citep{yin2025videoarm} and MM-mem~\citep{lian2026mm-mem} pioneered hierarchical memory architectures. However, they maintain these hierarchies as edge-less perceptual buffers or rely on passive, uncertainty-driven vertical drill-down. \colorours bridges this gap by coupling a three-tier hierarchical abstraction with a causal graph topology, encoding multi-granular semantics and localized details to ground subsequent retrieval. 
Furthermore, to decouple perception and reasoning for long videos, we construct a purely textual memory. Unlike previous leading methods~\citep{zhang2026dvd,yin2025videoarm} that require additional perception models to revisit raw video frames during retrieval
, our approach relies solely on interactions with text memory.

\subsection{Agentic Retrieval Mechanism}
Early information-seeking approaches~\citep{luo2026videorag,jeong2025videorag2} rely on naive Retrieval-Augmented Generation~\citep{lewis2020rag1,gao2023rag2}, using static dense embeddings to recall contexts. However, semantic similarity does not guarantee logical relevance. A visually similar clip may lack causal connection to the query, and these single-turn retrievals cannot dynamically self-correct. To address this, pioneering works like VideoAgent~\citep{wang2024videoagent} and DVD~\citep{zhang2026dvd} introduced agentic retrieval, enabling VLMs to actively search. Yet, constrained by flat clip databases, their exploration often degrades into blind trial-and-error. \colorours extends the agentic loop by coupling it with our graph memory. Through a tool-augmented Observation-Reason-Action loop, we unleash the agent's reasoning capabilities to explore the memory across multiple dimensions and locate relevant information.
\begin{figure*}[t]
  \includegraphics[width=\linewidth]{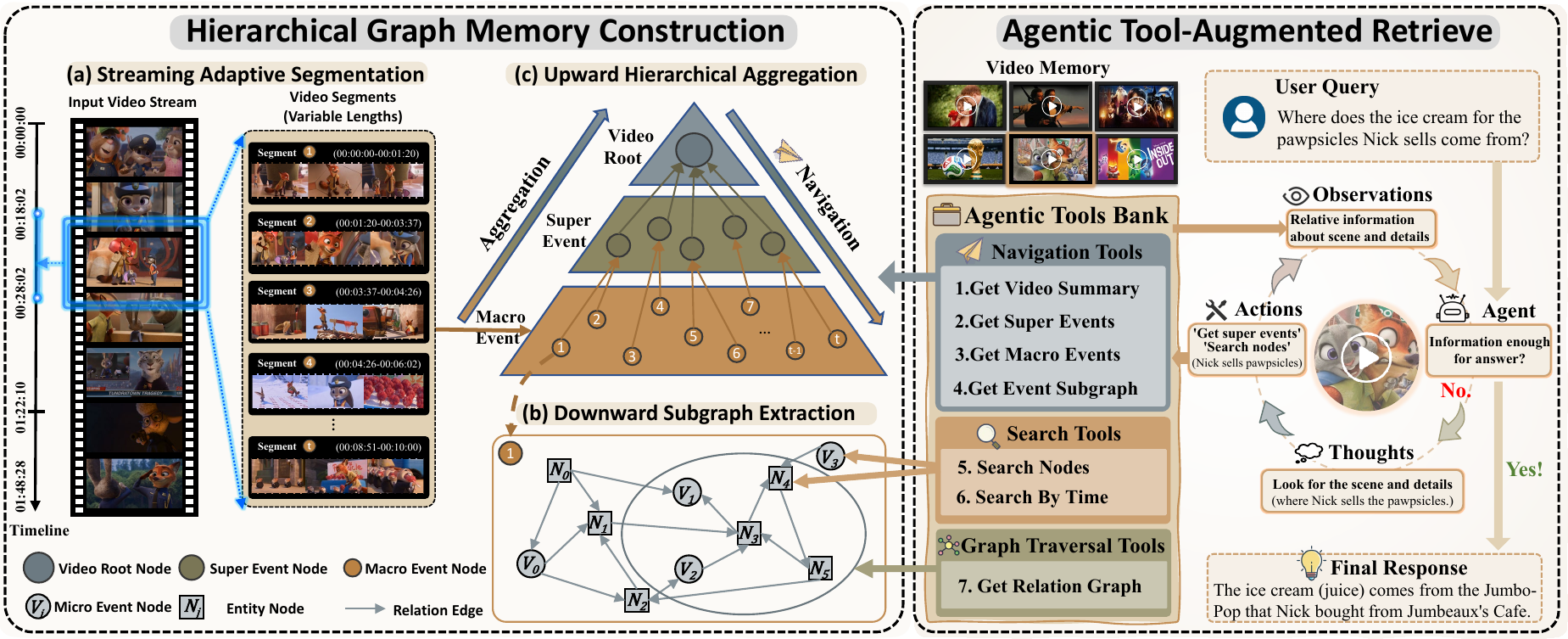}
  \caption{
    \textbf{Architectural workflow of \colorours.} (Left) Memory construction comprises three phases: streaming adaptive segmentation, downward subgraph extraction and upward hierarchical aggregation. (Right) Three tool categories support an agentic retrieval mechanism driven by an Observation-Reason-Action loop.
  }
\end{figure*}
\section{Method}

We formulate long-video understanding as a decoupled paradigm comprising two stages: persistent memory construction and tool-augmented agentic retrieval. Given a video stream $V$, a perception model $\mathcal{P}$ first processes the visual inputs in a streaming fashion to construct a structured, purely textual Hierarchical Graph Memory, denoted as $\mathcal{G}$. Subsequently, upon receiving a text query $Q$, a reasoning model $\mathcal{R}$ equipped with a tool bank $\mathcal{T}$ actively explores $\mathcal{G}$ through a multi-step Observation-Reason-Action loop. During inference, $\mathcal{R}$ operates only on this pure text memory, never raw video information. This iterative exploration extracts a concise set of task-relevant clues $C$, yielding the final answer $A = \mathcal{R}(Q, C)$.

\subsection{Hierarchical Graph Memory}

\colorours conceptualizes a long video stream as a coarse-to-fine three-tier semantic topology comprising the \textit{Video Root}, \textit{Super Events} and \textit{Macro Events} layers. Beneath the foundational \textit{Macro Events} layer, we formulate a local subgraph to characterize entities, micro-events, and their spatiotemporal and causal dependencies. To construct this memory representation, the pipeline first adaptively \textbf{segments the streaming video} to yield semantically self-contained \textit{Macro Events} as anchors. subsequently, the system \textbf{extracts fine-grained local subgraph} structures downward to preserve visual details, while \textbf{aggregating nodes upward} to distill macro-level hierarchical semantics.

\paragraph{Streaming Adaptive Segmentation.} 
We introduce a streaming adaptive segmentation mechanism driven by semantic boundaries, rather than relying on rigid, fixed-length chunking (e.g., 30-second intervals) common in prior works~\citep{zhang2026dvd}. Specifically, at each iteration $k$, the system maintains a temporal sliding window $W_k = [t_{\text{start}}^k, t_{\text{start}}^k + \tau]$, where $t_{\text{start}}^k$ denotes the initial timestamp of the current window and $\tau$ represents the maximum window duration. Within $W_k$, the perception model $\mathcal{P}$ executes temporal grounding to resolve a localized set of complete \textit{Macro Events}, denoted as $\{e_1, e_2, \dots, e_N\}$. The concluding boundary of this final event $e_N$ directly serves as the starting point for the subsequent window. This paradigm yields two benefits: first, the adaptive semantic partitioning ensures that each \textit{Macro Event} node remains semantically intact and self-contained, thereby avoiding the arbitrary truncation inherent in previous fixed-window strategies. Second, it caps the maximum video length of a single input to $\mathcal{P}$ within the horizon $\tau$, bounding context window pressure during the perception phase.

\begin{table*}[t]
\centering
\footnotesize
\setlength{\tabcolsep}{6pt}
\renewcommand{\arraystretch}{1.08}
\begin{tabularx}{\textwidth}{@{} l l X @{}}
\toprule
\textbf{Symbol} & \textbf{Type} & \textbf{Definition} \\
\midrule

\rowcolor{gray!10}[0pt][0pt]
\multicolumn{3}{@{}l@{}}{\textit{\textbf{Hierarchy nodes (one per video)}}} \\
\addlinespace[2pt]
$v^{R}$    & Video Root    & Global root: \texttt{title}, \texttt{description}, \texttt{themes}, \texttt{key\_entities}, \texttt{time\_range}. \\
$v^{S}$    & Super Event   & Narrative-phase: \texttt{label}, \texttt{description}, \texttt{time\_range}, $\{v^{Mc}\}$ children, deduplicated \texttt{key\_entities}. \\
$v^{Mc}$   & Macro Event   & Per-episode summary: \texttt{label}, \texttt{summary}, \texttt{time\_range}, child micro-event ids, top-$k$ \texttt{key\_entities}. \\
\midrule

\rowcolor{gray!10}[0pt][0pt]
\multicolumn{3}{@{}l@{}}{\textit{\textbf{Subgraph nodes $\mathcal{V}_i = \mathcal{V}_i^{E}\cup\mathcal{V}_i^{M}$ (one subgraph per Macro Event)}}} \\
\addlinespace[2pt]
$\mathcal{V}_i^{E}$ & Entity         & Persistent referent: \texttt{name}, \texttt{entity\_type} $\in\{$\textsc{Person}, \textsc{Object}, \textsc{Location}, \textsc{Group}$\}$, structured \texttt{attributes}, \texttt{description}. \\
$\mathcal{V}_i^{M}$ & Micro-event    & Atomic action with explicit temporal extent: \texttt{time\_range}, \texttt{subject}, \texttt{object}, \texttt{action}, \texttt{description}. \\
\midrule

\rowcolor{gray!10}[0pt][0pt]
\multicolumn{3}{@{}l@{}}{\textit{\textbf{Subgraph edges $\mathcal{E}_i$, three topological categories}}} \\
\addlinespace[2pt]
\textit{Spatial-attribute} & $\mathcal{V}_i^{E}\!\leftrightarrow\!\mathcal{V}_i^{E}$ & \texttt{LOCATED\_IN}, \texttt{NEXT\_TO}, \texttt{ATTACHED\_TO}, \texttt{PART\_OF}, \texttt{IS\_SAME\_AS}, \texttt{ANNOTATES}, static layout, composition, coreference, and on-screen attribute. \\
\textit{Subject-object}    & $\mathcal{V}_i^{E}\!\leftrightarrow\!\mathcal{V}_i^{M}$ & \texttt{PERFORMS}, \texttt{RECEIVES}, \texttt{USES\_TOOL}, \texttt{HAPPENS\_IN}, \texttt{CONTEXT\_FOR}, role assignment of entities to micro-events and contextual annotation of events. \\
\textit{Temporal-causal}   & $\mathcal{V}_i^{M}\!\to\!\mathcal{V}_i^{M}$ & \texttt{BEFORE}, \texttt{OVERLAP}, \texttt{CAUSES}, \texttt{PREVENTS}, \texttt{SUBEVENT\_OF}, chronology and causality among micro-events. \\
\midrule

\rowcolor{gray!10}[0pt][0pt]
\multicolumn{3}{@{}l@{}}{\textit{\textbf{Cross-tier edges $\mathcal{E}^{H}$}}} \\
\addlinespace[2pt]
\textit{Hierarchical}      & $v^{Mc}\!\to\!v^{S}$, $\;v^{S}\!\to\!v^{R}$ & \texttt{SUBEVENT\_OF} edges that wire the three-tier topology. \\
\textit{Macro causal / progression} & $v^{Mc}\!\to\!v^{Mc}$, $\;v^{S}\!\to\!v^{S}$ & Inter-episode reasoning: \texttt{CAUSES}, \texttt{PREVENTS}, \texttt{ENABLES} (intra-Super), \texttt{LEADS\_TO}, \texttt{RESOLVES}, \texttt{CONTRASTS\_WITH} (cross-Super). \texttt{BEFORE} edges between consecutive Super Events are auto-generated. \\
\bottomrule
\end{tabularx}
\caption{Schema of the Hierarchical Graph Memory $\mathcal{G}$. The hierarchy spans three tiers ($v^{R}$, $v^{S}$, $v^{Mc}$, and the Macro Event leaves whose contents are unfolded into local subgraphs $g_i$). Each subgraph factorises a Macro Event into entities, micro-events, connected by three edge categories that decouple layout, action role binding, and temporal-causal flow.}
\label{tab:graph-def}
\end{table*}
\paragraph{Downward Subgraph Extraction.} 
For each segmented \textit{Macro Events} $e_i$, we construct a local spatiotemporal subgraph $g_i = (\mathcal{V}_i, \mathcal{E}_i)$ to capture fine-grained visual details. 
While conventional LLM-based graph memories~\citep{edge2024graphrag} encapsulate an entire event into a single static $\langle$\textit{Subject entity}, \textit{Relation edge}, \textit{Object entity}$\rangle$ triplet, they struggle to model the highly dynamic temporal evolution and causal chains inherent in long videos. To address this limitation, we introduce a novel \textit{Video-centric Graph Ontology} where the vertex set is formally defined as $\mathcal{V}_i = \mathcal{V}_i^E \cup \mathcal{V}_i^M$, incorporating micro-events ($\mathcal{V}_i^M$) as distinct nodes alongside standard entities ($\mathcal{V}_i^E$). These vertices are interconnected via a heterogeneous edge set $\mathcal{E}_i$ comprising three distinct topological relations: \textit{spatial-attribute edges} ($\mathcal{V}_i^E \leftrightarrow \mathcal{V}_i^E$) delineate the relative spatial layouts and static physical attributes among entities, \textit{subject-object edges} ($\mathcal{V}_i^E \leftrightarrow \mathcal{V}_i^M$) anchor discrete entities to the actions they initiate or undergo, and directed \textit{temporal-causal edges} ($\mathcal{V}_i^M \rightarrow \mathcal{V}_i^M$) model the chronological progression and causal dependencies between micro-events. We use the perception model $\mathcal{P}$ to parse the video clip to extract this structured subgraph $g_i$ alongside a textual summary of the \textit{Macro Event}.

\paragraph{Upward Hierarchical Aggregation.} 
Upon establishing the foundational graph layer, the system executes a bottom-up semantic abstraction to provide a global navigation backbone for subsequent retrieval. Specifically, the localized textual descriptions of all \textit{Macro Events} are fed into the model $\mathcal{P}$ as the foundational leaf nodes of the three-tier topology. Guided by temporal adjacency and semantic affinity, $\mathcal{P}$ clusters and distills neighboring, highly correlated \textit{Macro Events} into \textit{Super Events} that span broader temporal scales, until ultimately converging at a single \textit{Video Root} node. This coarse-to-fine topology structurally spans from micro-level actions to the overall narrative, empowering the retrieval agent with efficient global navigation and precise target localization.

\begin{table*}[t]
\centering
\footnotesize
\setlength{\tabcolsep}{6pt}
\renewcommand{\arraystretch}{1.1}
\begin{tabularx}{\textwidth}{@{} l l X X @{}}
\toprule
\textbf{Tool} & \textbf{Parameters} & \textbf{Output} & \textbf{Purpose} \\
\midrule

\rowcolor{gray!10}[0pt][0pt]
\multicolumn{4}{@{}l@{}}{\textit{\textbf{Hierarchical Navigation}}} \\
\addlinespace[2pt]
\texttt{GetSummary}        & -- & Video-level summary & Retrieve the overview of video. \\
\texttt{GetSuperEvent}     & \texttt{super\_id} & Super-event records: \texttt{label}, \texttt{time\_range}, and child-macro. & Browse high-level narrative segments. \\
\texttt{GetMacroEvent}     & \texttt{super\_id} \emph{or} \texttt{macro\_ids} & Macro-event records: \texttt{label}, \texttt{time\_range}, and \texttt{key\_entities}. & Drill from a super-event into its constituent macro-events. \\
\texttt{GetSubgraph}       & \texttt{macro\_id} & Local subgraph of one macro-event: entities, micro-events, on-screen text, and key edges. & Inspect fine-grained dynamics inside a single macro-event. \\
\midrule

\rowcolor{gray!10}[0pt][0pt]
\multicolumn{4}{@{}l@{}}{\textit{\textbf{Precise Search}}} \\
\addlinespace[2pt]
\texttt{SearchNodes}       & \texttt{query}, \texttt{top\_k}, \texttt{node\_types} & Top-$k$ nodes ranked by embedding similarity, each annotated with its parent macro-event and one-hop ego-graph context. & Dense semantic retrieval over entities, micro-events, and on-screen text within a unified embedding space. \\
\texttt{SearchByTime}      & \texttt{start\_sec}, \texttt{end\_sec} & Macro-events whose time intervals overlap $[t_s, t_e]$, paired with their parent super-event labels. & Time-anchored localization given a temporal range. \\
\midrule

\rowcolor{gray!10}[0pt][0pt]
\multicolumn{4}{@{}l@{}}{\textit{\textbf{Graph Traversal}}} \\
\addlinespace[2pt]
\texttt{GetRelationGraph}  & \texttt{node\_id} & Outgoing and incoming edges of the target node, with relation types, labels, and rationales. & Multi-hop traversal along spatial-attribute, subject-object, and temporal-causal edges. \\
\bottomrule
\end{tabularx}
\caption{Specification of the seven tools available to the reasoning model $\mathcal{R}$ during the Observation-Reason-Action loop. The toolkit is organized into three categories aligned with the agent's retrieval intent: navigating the three-tier hierarchy, executing semantic or temporal search, and traversing local graph neighborhoods.}
\label{tab:tool-spec}
\end{table*}
\subsection{Agentic Tool-Augmented Retrieval}
We introduce a tool-augmented agentic retrieval mechanism to exploit our structured, multi-granular graph memory. Equipping the reasoning model $\mathcal{R}$ with a graph-access toolbox across three functional dimensions and an Observation-Reason-Action loop, we transform static video understanding into a multi-step, proactive reasoning process.

\paragraph{Multi-Dimensional Tool Bank.} 
To facilitate deep retrieval and reasoning over the hierarchical graph memory, we construct a graph-access toolbox encompassing three functional dimensions. First, the \textbf{\textit{hierarchical navigation tools}}, consisting of $\textit{GetSummary}(\cdot)$, $\textit{GetSuperEvent}(\cdot)$, $\textit{GetMacroEvent}(\cdot)$, and $\textit{GetSubgraph}(\cdot)$, allow the model $\mathcal{R}$ to progressively unpack the three-tier topology from a top-down, coarse-to-fine perspective, traversing from global narratives down to super-events, macro-events and localized subgraphs at varying granularities. Second, the \textbf{\textit{precise search tools}}, including $\textit{SearchNodes}(\cdot)$ and $\textit{SearchByTime}(\cdot)$, support vector-based semantic retrieval via text embeddings and chronological filtering within targeted temporal windows, enabling rapid localization of relevant nodes. Third, the \textbf{\textit{local graph traversal tool}} $\textit{GetRelationGraph}(\cdot)$ extracts the topological neighborhood of a target node, allowing the model to execute multi-hop tracing along spatial, temporal, or causal edges to resolve complex cross-spatiotemporal logic chains. Collectively, these three tool categories formulate an expressive action space, providing the retrieval agent with the capability to shift its perspective between the macroscopic semantic hierarchy and the underlying dense graph details.

\paragraph{Agentic Observation-Reason-Action Loop.} 
Our retrieval mechanism operates within an Observation-Reason-Action loop driven by the model $\mathcal{R}$. At each step $t$, the model evaluates the initial user query $Q$ alongside the historical execution trajectory $\mathcal{H}_{t-1}$ to determine the proper tool execution and corresponding parameters $a_t$:\begin{equation}
a_t = \mathcal{R}(Q, \mathcal{H}_{t-1})
\end{equation}
Executing the action $a_t$ invokes the designated tool from our toolkit to yield a environmental observation $o_t$ from the graph memory $\mathcal{G}$. To prevent multi-turn interactions from accumulating irrelevant textual noise, which induces the ``lost in the middle'' phenomenon within long contexts, we eschew direct concatenation of raw observational outputs. Instead, the model $\mathcal{R}$ distills $o_t$ to extract only the task-relevant evidence cues conditional on the query, denoted as 
\begin{equation}
c_t = \mathcal{R}(o_t, Q). 
\end{equation}
Subsequently, the current action-clue pair is appended to the historical trajectory to transition the working memory state of the agent:
\begin{equation}
\mathcal{H}_t = \mathcal{H}_{t-1} \cup \{(a_t, c_t)\}
\end{equation}
By obtaining raw observations and aggregating only relevant clues, this mechanism guarantees that the reasoning model $\mathcal{R}$  exclusively attends to high-quality evidence. Through this entire Agentic Loop, the model  $\mathcal{R}$ can fully explore the memory space to gather the essential evidence required for the final response.
\definecolor{rowgray1}{HTML}{E5DFD9} 
\definecolor{rowgray2}{HTML}{E3DFE1} 
\definecolor{rowgray3}{HTML}{DFE3DE} 
\definecolor{rowgray4}{HTML}{DEE0E5}
\definecolor{rowgray5}{HTML}{FFFFFF}

\begin{table*}[t]
\centering
\renewcommand{\arraystretch}{1}
\renewcommand{\arraystretch}{1.2} 
\resizebox{\textwidth}{!}{
\begin{tabular}{ll ccccccc cccc}
\toprule
\multirow{2}{*}{\textbf{Methods}} & \multirow{2}{*}{\textbf{Reason Model}} & \multicolumn{7}{c}{\textbf{LVBench}} & \multicolumn{2}{c}{\textbf{LongVideoBench}} & \textbf{Video-MME} & \textbf{EgoSchema} \\
\cmidrule(lr){3-9} \cmidrule(lr){10-11} \cmidrule(lr){12-12} \cmidrule(lr){13-13}
& & \textbf{Avg.$\uparrow$} & \textbf{ER$\uparrow$} & \textbf{EU$\uparrow$} & \textbf{Rea$\uparrow$} & \textbf{KIR$\uparrow$} & \textbf{Sum$\uparrow$} & \textbf{TG$\uparrow$} & \textbf{Avg.$\uparrow$} & \textbf{Long$\uparrow$} & \textbf{Long(w/o sub)$\uparrow$} & \textbf{Val$\uparrow$} \\
\midrule
\rowcolor{rowgray1} \multicolumn{13}{l}{\textit{\textbf{End-to-End VLMs (Proprietary)}}} \\
GPT-4o & GPT-4o & 48.9 & 48.9 & 49.5 & 50.3 & 48.1 & 50.0 & 40.9 & 66.7 & 60.9 & 65.3 & 70.4 \\
Gemini-2.0-Flash & Gemini-2.0-Flash & 48.6 & 47.4 & 48.5 & 44.4 & 56.8 & 41.4 & 39.3 & $-$ & 45.7 & 63.0 & 71.2 \\
Gemini-2.5-Pro & Gemini-2.5-Pro & 72.0 & 71.5 & 71.1 & 67.7 & 80.0 & 63.5 & 69.1 & 71.0 & 68.6 & 75.9 & 72.8 \\
Gemini-3.1-Pro & Gemini-3.1-Pro & 78.2 & 78.1 & 76.7 & 74.1 & 86.3 & 70.7 & 81.4 & 78.6 & 77.0 & 80.3 & 76.4 \\
OpenAI-o3 & OpenAI-o3 & 57.1 & 57.6 & 56.4 & 50.8 & 62.9 & 67.2 & 46.8 & 66.7 & 60.6 & 64.7 & 63.2 \\
Seed1.5VL & Seed1.5VL-Thinking & 64.6 & 65.4 & 63.4 & 68.0 & 53.6 & 63.7 & 46.6 & $-$ & 74.4 & $-$ & $-$ \\
\midrule
\rowcolor{rowgray2} \multicolumn{13}{l}{\textit{\textbf{End-to-End VLMs (Open-Source)}}} \\
Qwen3-VL & Qwen3-VL-235B-A22B-Thinking & 63.6 & 63.7 & 62.6 & 63.1 & 62.6 & 65.5 & 59.6 & 71.4 & 67.1 & 71.2 & 75.9 \\
Qwen3.5 & Qwen3.5-35A3B & 71.3 & 72.2 & 69.4 & 74.4 & 68.9 & 67.5 & 66.2 & $-$ & $-$ & 62.6 & $-$ \\
GLM-4.6V & GLM-4.6V-106B-A12B & 59.5 & 58.8 & 59.2 & 69.7 & 57.8 & 61.8 & 53.9 & 67.6 & 58.7 & 66.0 & 68.8 \\
GLM-4.5V & GLM-4.5V-106B-A12B & 53.4 & 55.3 & 53.1 & 57.5 & 55.3 & 39.6 & 49.9 & 66.4 & 54.1 & 64.7 & 68.4 \\
InternVL2.5 & InternVL2.5-78B & 43.6 & 43.8 & 42.0 & 51.0 & 37.9 & 42.1 & 36.8 & $-$ & $-$ & 62.6 & $-$ \\
InternVL3.5 & InternVL3.5-30BA3B & 44.4 & 42.7 & 44.1 & 48.3 & 46.4 & 36.2 & 40.9 & 62.9 & 52.7 & 64.1 & 86.8 \\
AdaRETAKE & AdaRETAKE & 53.3 & 53.0 & 50.7 & 54.7 & 62.2 & 37.9 & 45.5 & 67.0 & $-$ & 65.0 & $-$ \\
\midrule
\rowcolor{rowgray3} \multicolumn{13}{l}{\textit{\textbf{Agentic Video Understanding Methods}}} \\
VideoTree & VideoTree & 28.8 & 30.3 & 25.1 & 31.9 & 26.5 & 25.5 & 27.7 & $-$ & $-$ & $-$ & 67.0 \\
VideoAgent & VideoAgent & 29.3 & 28.0 & 30.3 & 28.0 & 28.0 & 36.4 & 29.3 & $-$ & $-$ & $-$ & 63.2 \\
VCA & VCA & 41.3 & 43.7 & 40.7 & 46.2 & 37.8 & 27.3 & 38.0 & $-$ & $-$ & $-$ & 73.6 \\
MR.~Video & MR.~Video & 60.8 & 59.8 & 57.4 & 57.7 & 71.4 & 50.0 & 58.8 & $-$ & 61.6 & 61.8 & 73.0 \\
M3-Agent & M3-Agent & 49.3 & $-$ & $-$ & $-$ & $-$ & $-$ & $-$ & $-$ & $-$ & 61.8 & $-$ \\
WorldMM-GPT & GPT-5 & 61.9 & $-$ & $-$ & $-$ & $-$ & $-$ & $-$ & $-$ & $-$ & 76.6 & $-$ \\
MM-Mem & MM-Mem & $-$ & $-$ & $-$ & $-$ & $-$ & $-$ & $-$ & $-$ & $-$ & 66.1 & $-$ \\
DVD & OpenAI-o3 & 74.2 & 73.4 & 73.3 & 70.7 & 80.4 & 74.1 & 72.3 & 71.6 & 68.6 & 67.3 & 76.6 \\
VideoARM & OpenAI-o3 & 79.7 & $-$ & $-$ & $-$ & $-$ & $-$ & $-$ & 78.0 & 76.4 & 81.2 & 76.2 \\
VideoSeek & GPT-5 & 68.4 & $-$ & $-$ & $-$ & $-$ & $-$ & $-$ & $-$ & 73.5 & 60.9 & $-$ \\
\midrule
\rowcolor{rowgray4} \multicolumn{13}{l}{\textit{\textbf{\ours (Ours, Plug-and-Play Agentic Framework)}}} \\
\rowcolor{rowgray5} \colorours & Qwen3-VL-235B-A22B-Thinking & \textbf{84.8} \up{21.2} & 84.6 & 85.2 & 80.6 & 85.6 & 84.5 & 87.7 & \textbf{86.3} \up{14.9} & \textbf{83.2} \up{16.1} & \textbf{86.2} \up{15.0} & \textbf{87.4} \up{11.5} \\
\rowcolor{rowgray5} \colorours & Gemini-2.5-Pro & \textbf{80.7} \up{8.7} & 77.1 & 83.6 & 81.1 & 82.5 & 91.4 & 86.4 & \textbf{78.6} \up{7.6} & \textbf{78.7} \up{10.1} & \textbf{85.0} \up{9.1} & \textbf{88.2} \up{15.4} \\
\rowcolor{rowgray5} \colorours & Gemini-3.1-Pro & \textbf{90.7} \up{12.5} & 90.1 & 90.6 & 89.6 & 91.4 & 89.7 & 91.8 & \textbf{92.9} \up{14.3} & \textbf{91.0} \up{14.0} & \textbf{92.1} \up{11.8} & \textbf{87.8} \up{11.4} \\
\midrule
\midrule
\textit{Human Expert} & \textit{Human} & \textit{94.4} & \textcolor{lightgray}{$-$} & \textcolor{lightgray}{$-$} & \textcolor{lightgray}{$-$} & \textcolor{lightgray}{$-$} & \textcolor{lightgray}{$-$} & \textcolor{lightgray}{$-$} & \textcolor{lightgray}{$-$} & \textcolor{lightgray}{$-$} & \textcolor{lightgray}{$-$} & \textcolor{lightgray}{$-$} \\
\bottomrule
\end{tabular}
}
\caption{\textbf{Main results on four long-video understanding benchmarks.} \colorours achieves state-of-the-art across all four benchmarks, with substantial gains over end-to-end baselines (\textcolor{green!60!black}{green}: improvement over the strongest end-to-end baseline using the same backbone).}
\label{tab:main-results}
\end{table*}

\section{Experiments}
\subsection{Experiment Setup}

\paragraph{Evaluation Benchmarks.} 
We use four challenging benchmarks: LVBench~\citep{lvbench} comprises 103 long videos of 30 mins to 2 hours and 1,549 QA pairs across six fine-grained dimensions. The LongVideoBench~\citep{longvideobench} validation set includes 753 videos and 1,337 questions, featuring 188 videos of 15 to 60 minutes. The long-video split of Video-MME~\citep{videomme} covers 300 videos (30 to 60 minutes) and 900 questions. EgoSchema~\citep{egoschema} contains egocentric clips focusing on reasoning. Together, these benchmarks span diverse video durations, task granularities, and reasoning depths.

\paragraph{Implementation Details.} 
In the memory construction phase, the continuous video stream is segmented via a $\tau = 10 min$ sliding window, and Gemini-3.1-Pro serves as the perception model to construct the hierarchical graph memory using raw video streams without any external transcripts. In the online reasoning phase, we evaluate Gemini-2.5-Pro~\citep{gemini2.5}, Gemini-3.1-Pro~\citep{deepmind_gemini}, and the open-source Qwen3-VL-235B-A22B-Thinking~\citep{bai2025qwen3vl} as the reasoning engines. We adopt Qwen3-Embedding~\citep{zhang2025qwen3embedding} for semantic vector computing and cap the agent's maximum tool-call budget at 12 steps. 

\paragraph{Baselines.} 
We benchmark \colorours against two categories of long-video systems:
(1) \textit{Vanilla Long Video VLMs}: we compare it directly against its underlying reasoning engines, covering top closed-source frontiers~\citep{gpt-4o,deepmind_gemini,guo2025seed1.5vl} and representative open-source~\citep{bai2025qwen3vl,qwen3_5_blog,GLM4.6V,zeng2025glm4.5,chen2024Internvl2.5,wang2025internvl3.5,wang2025adaretake}.
(2) \textit{Memory-based Video LLMs}: we evaluate our framework against established long-form video memory baselines~\citep{wang2025videotree,wang2024videoagent,yang2025vca,pang2025mr.video,long2025m3agent,yeo2025worldmm,lian2026mm-mem} alongside the latest SOTA memory-driven systems~\citep{zhang2026dvd,yin2025videoarm,lin2026videoseek}.

\subsection{Main Results}

\paragraph{SOTA Performance Across Diverse Benchmarks.} 
Table~\ref{tab:main-results} summarizes the performance of \colorours across four benchmarks. Results demonstrate that \colorours establishes new SOTA across all evaluation metrics. Specifically, on LVBench, \colorours achieves a peak score of 90.7 with 12.5 points improvement over the strongest native closed-source baseline Gemini-3.1-Pro, significantly narrowing the gap to human expert performance to a 3.7 points. On LongVideoBench and Video-MME, our framework attains 92.9 and 92.1, yielding substantial improvements of 14.3 and 11.8 points. Furthermore, even on EgoSchema, which features egocentric perspectives, \colorours secures a top score of 88.2. These consistent improvements across both open-source and closed-source reasoning engines validate the generality and pronounced superiority of our plug-and-play architectural paradigm.

\begin{table}[h]
\centering
\footnotesize
\setlength{\tabcolsep}{4pt}       
\renewcommand{\arraystretch}{0.92} 

\resizebox{0.65\linewidth}{!}{
\begin{tabular}{@{} l l c c @{}}
\toprule
\textbf{Role} & \textbf{Model} & \textbf{Context Windows} & \textbf{LVB} \\
\midrule

\multicolumn{4}{@{}l}{\textit{\textbf{Vanilla Full Video}}} \\[2pt] 
\multirow{3}{*}{End-to-End} 
 & Gemini-3.1-Pro             & 265K & 78.2 \\
 & Gemini-2.5-Pro             & 784K & 72.0 \\
 & Qwen3-VL-235B-A22B-Thinking           & 240K & 63.6 \\
\midrule

\multicolumn{4}{@{}l}{\textit{\textbf{MemDreamer (Ours)}}} \\[2pt] 
Perception & Gemini-3.1-Pro             & 40.3K & - \\
\cdashline{1-4} 
\addlinespace[2pt]  
\multirow{3}{*}{Reasoning} 
 & Gemini-3.1-Pro             & \textbf{6.2K} & \textbf{90.7} \\
 & Gemini-2.5-Pro             & \textbf{6.3K} & \textbf{80.7} \\
 & Qwen3-VL-235B-A22B-Thinking           & \textbf{5.9K} & \textbf{84.8} \\
\bottomrule
\end{tabular}
}
\caption{Comparison of input context length requirements and overall performance on LVBench (LVB).}
\label{tab:context_length_comparison}
\end{table}
\paragraph{Mitigating Context Overload via Decoupled Reasoning.} 
Table~\ref{tab:context_length_comparison} and Figure~\ref{fig:teaser_combined} compare the context length of our decoupled paradigm against conventional native end-to-end full-context ingestion. The vanilla approach forces the vision-language model to ingest massive sequences ranging from 240K to 784K tokens, severely degrading reasoning accuracy due to spatial-attention dilution and the ``lost in the middle'' effect. Conversely, by segregating perception from reasoning, \colorours avoids stacking redundant visual tokens. During agentic reasoning, our tool-augmented navigation operates over a condensed topology, restricting the active context to 5.9K--6.3K tokens, delivering an approximate 40$\times$ reduction versus full-video ingestion. Reducing token noise and redundancy unlocks a 21.2-point leap for Qwen3-VL from 63.6 to 84.8, empirically showing that structured agentic reasoning is a superior alternative to brute-force token scaling.

\subsection{Analysis}
\begin{table}[h]
\centering
\renewcommand{\arraystretch}{0.98} % 收紧行高
\resizebox{0.75\linewidth}{!}{
\begin{tabular}{lccc}
\toprule
\multirow{2}{*}{\textbf{Model}} & \multirow{2}{*}{\textbf{AIME2025}} & \multicolumn{2}{c}{\textbf{LVBench}} \\
\cmidrule{3-4}
 & & \textbf{End-to-end} & \textbf{\colorours} \\
\midrule
Claude-Opus-4.6            & \textbf{99.8} & 67.9 & 85.1 \\
Gemini-3.1-Pro             & 98.1 & \textbf{78.2} & \textbf{90.7} \\
Kimi-K2.5                    & 96.1 & 75.9 & 79.2 \\
GPT-5                      & 94.6 & 60.1 & 84.4 \\
Qwen3-VL-235B-A22B-Thinking& 89.7 & 63.6 & 84.8 \\
Gemini-2.5-Pro             & 88.0 & 72.0 & 80.7 \\
Qwen3-VL-235B-A22B-Instruct& 74.7 & 67.7 & 71.5 \\
gpt-4o-2024-11-20          & 61.9 & 48.9 & 68.5 \\
\midrule
\multicolumn{4}{l}{\textit{Statistical Correlation with Agentic Capability}} \\
\midrule
Pearson $R$ $\uparrow$     & -     & 0.702  & \textbf{0.897} \\
$p$-value $\downarrow$     & -     & 0.052  & $\bm{<}$\textbf{0.01} \\
\bottomrule
\end{tabular}
}
\caption{Correlation between Agentic reasoning capabilities (AIME 2025) and long video understanding.}
\label{tab:agentic_correlation_tall}
\end{table}
\paragraph{Paradigm Shift to Pure Agentic Reasoning.} 
Table~\ref{tab:agentic_correlation_tall} uncovers the correlation between an LLM's logical capacity and long-video understanding performance. We evaluate eight models~\citep{team2026kimi-k2.5,anthropic2024claudeopus} on the widely used AIME2025 reasoning benchmark and compute the Pearson correlation ($R$) against their LVBench scores. Under the traditional end-to-end paradigm, the correlation is restricted to 0.702 with an insignificant $p$-value of 0.052, demonstrating that raw video ingestion creates perceptual barriers that hinder models from leveraging their reasoning capacity. Conversely, integrating \colorours scales this correlation to \textbf{0.897} with statistical significance ($p < 0.01$). This establishes a paradigm shift: supported by structured graph memory and agentic retrieval mechanism, long-video understanding bypasses the context-window limits, scales with the model's logical reasoning capability.

\begin{table}[h]
\centering
\renewcommand{\arraystretch}{0.92}
\resizebox{0.7\linewidth}{!}{
\begin{tabular}{ll c}
\toprule
\textbf{Perception Model} & \textbf{Reasoning Model} & \textbf{LVBench} \\
\midrule
\multirow{3}{*}{Gemini-2.5-Pro} 
 & Gemini-2.5-Pro             & 78.2 \\
 & Gemini-3.1-Pro             & 90.3 \\
 & Qwen3-VL-235B-A22B-Thinking   & 83.4 \\
\midrule
\multirow{3}{*}{Gemini-3.1-Pro} 
 & Gemini-2.5-Pro             & 80.7 \\
 & Gemini-3.1-Pro             & 90.7 \\
 & Qwen3-VL-235B-A22B-Thinking   & 84.8 \\
\bottomrule
\end{tabular}
}
\caption{Ablation study on different base models.}
\label{tab:ablation_overall_only}
\end{table}
\paragraph{Robustness via Decoupled Base Models.} 
Table~\ref{tab:ablation_overall_only} presents the ablation under varying combinations of perception and reasoning backbones. Because \colorours only requires the perception model to process localized short-form video clips (under 10 min), the capability gaps between different perception backbones are largely smoothed out. When driving the same downstream reasoning engine, either Gemini-3.1-Pro or Qwen3-VL, the final performance fluctuations caused by swapping the underlying perception model are merely 0.4 and 1.4 percentage points, respectively. This demonstrates that our decoupled framework exhibits high perceptual error tolerance, effectively alleviating the over-reliance on expensive, long-context perception for macroscopic video understanding.

\begin{figure}[h]
\centering
\includegraphics[width=0.8\linewidth]{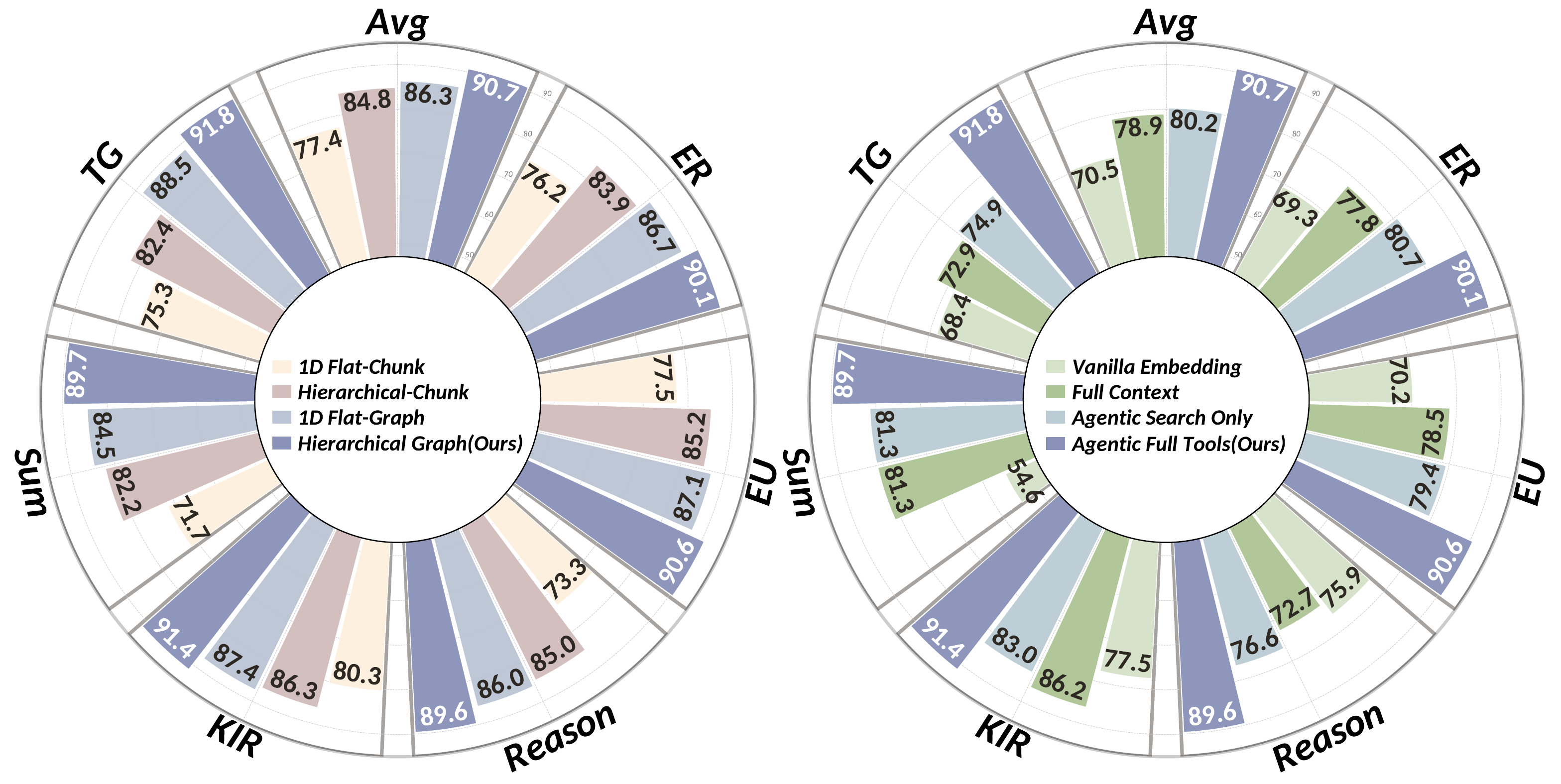}\\[0.6ex]
{\footnotesize
\begin{tabular}{@{}p{0.4\linewidth}@{}p{0.4\linewidth}@{}}
\centering (a) Memory Construction & \centering\arraybackslash (b) Retrieval Mechanism
\end{tabular}}
\caption{Ablations of different memory designs and retrieval strategies across LVBench sub-categories.}
\label{fig:radar_chart}
\end{figure}

\begin{table}[h]
\begin{minipage}[t]{0.48\textwidth}
\centering
\small
\renewcommand{\arraystretch}{0.92}
\setlength{\tabcolsep}{3pt}
\begin{tabular}{@{} l c c c @{}}
\toprule
\multirow{2}{*}{\textbf{Memory Architecture}} & \multicolumn{2}{c}{\textbf{Components}} & \multirow{2}{*}{\textbf{LVBench}} \\
\cmidrule(lr){2-3}
& \textbf{Hierarchy} & \textbf{Graph} & \\
\midrule
Flat-Chunk            & \ding{55} & \ding{55} & 77.4 \\
Flat-Graph            & \ding{55} & \ding{51} & 84.8 \\
Hierarchical-Chunk    & \ding{51} & \ding{55} & 86.3 \\
\midrule
\multicolumn{1}{@{} >{\columncolor{rowgray4}[0pt][3pt]}l}{\textbf{Hierarchical-Graph} ({\scriptsize \textbf{Ours}})} & 
\multicolumn{1}{>{\columncolor{rowgray4}[3pt][3pt]}c}{\ding{51}} & 
\multicolumn{1}{>{\columncolor{rowgray4}[3pt][3pt]}c}{\ding{51}} & 
\multicolumn{1}{>{\columncolor{rowgray4}[3pt][0pt]}c @{}}{\textbf{90.7}} \\
\bottomrule
\end{tabular}
\caption{Ablation study on the memory design.}
\label{tab:memory_ablation}
\end{minipage}
\hfill
\begin{minipage}[t]{0.50\textwidth}
\centering
\small
\setlength{\tabcolsep}{10pt}
\begin{tabularx}{\columnwidth}{@{} >{\raggedright\arraybackslash}X c c c @{}}
\toprule
\multirow{2}{*}{\textbf{Retrieval Strategy}} & \multicolumn{3}{c}{\textbf{LVBench}} \\
\cmidrule(lr){2-4}
& \textbf{Avg.} & \textbf{Sum.} & \textbf{Rea.} \\
\midrule
Vanilla Embedding Similarity & 70.5 & 54.6 & 75.9 \\
Full Memory Context          & 78.9 & 81.3 & 72.7 \\
Agentic Search Only          & 80.2 & 81.3 & 76.6 \\
\midrule
\multicolumn{1}{@{} >{\columncolor{rowgray4}[0pt][\tabcolsep]\raggedright\arraybackslash}X}{\textbf{Agentic Full Tools} ({\scriptsize \textbf{Ours}})} & 
\multicolumn{1}{>{\columncolor{rowgray4}[\tabcolsep][\tabcolsep]}c}{\textbf{90.7}} & 
\multicolumn{1}{>{\columncolor{rowgray4}[\tabcolsep][\tabcolsep]}c}{\textbf{89.7}} & 
\multicolumn{1}{>{\columncolor{rowgray4}[\tabcolsep][0pt]}c @{}}{\textbf{89.6}} \\
\bottomrule
\end{tabularx}
\caption{Ablation study on memory retrieval strategies.}
\label{tab:retrieval_strategy_ablation}
\end{minipage}
\end{table}

\paragraph{Hierarchy and graph contribute complementary gains.}
Table~\ref{tab:memory_ablation} and Figure~\ref{fig:radar_chart} isolates the independent and joint contributions of the hierarchical topology (\textit{Hierarchy}) and cross-node relational edges (\textit{Graph}). Removing both components \textit{1D Flat-Chunk} yields the worst baseline performance of 77.4. Introducing topological edges alone \textit{1D Flat-Graph} drives a prominent 7.4-point performance gain to 84.8, showing that temporal and causal graph connections effectively preserve event continuity. Similarly, incorporating the bottom-up abstraction layer alone \textit{Hierarchical-Chunk} raises the score to 86.3, validating that a multi-granular macro-skeleton prevents the retrieval agent from getting trapped in localized trivial details. Our full design \textit{Hierarchical-Graph} achieves the peak performance of 90.7. This compounding gain demonstrates a strong effect, where the macroscopic hierarchy and the underlying dense graph complement each other to optimize both global path navigation and localized target grounding.

\paragraph{Tool-augmented retrieval beats embedding
lookup.} Table~\ref{tab:retrieval_strategy_ablation} and Figure~\ref{fig:radar_chart} evaluates distinct memory access mechanisms to verify the necessity of our tool-augmented agentic retrieval. Our strategy \textit{Agentic Full Tools} achieves an score of 90.7, outperforming the traditional static match baseline \textit{Vanilla Embedding Similarity} by 20.2 points. This demonstrates that multi-step tool invocation captures significantly more precise cues than single-turn passive recall. While feeding the entire graph skeleton Full Memory Context benefits summarization with a score of 81.3, its performance on reasoning sharply degrades to 72.7. This
 confirms that passing the context window with unfiltered, graph structures introduces severe topological noise that disrupts critical logical deduction. Furthermore, restricting the agent to semantic retrieval alone Agentic Search Only yields a suboptimal accuracy of 80.2, demonstrating that the completeness of our multi-dimensional toolkit is critical for efficient long-form video understanding.

\begin{table}[h]
\centering
\footnotesize
\renewcommand{\arraystretch}{1.0}
\begin{tabular*}{0.75\linewidth}{@{\extracolsep{\fill}} c c c c @{}}
\midrule
\textbf{$T_{\max}$} & \textbf{LVBench} & \textbf{Avg.\ rounds} & \textbf{Tokens/round} \\
\midrule
8    & 88.7            & 2.87 & 5{,}990 \\
10   & 89.3            & 3.00 & 6{,}141 \\
12   & \textbf{90.7}   & 3.06 & 6{,}181 \\
15   & 90.2            & 3.07 & 6{,}222 \\
\midrule
\end{tabular*}
\caption{Ablation on the round budget $T_{\max}$ with $\mathcal{R}{=}$Gemini-3.1-Pro and the full toolkit. Avg.\ rounds is the mean number of rounds actually taken before $\mathcal{R}$ emits an answer. Tokens/round is the mean input length per tool-call round.}
\label{tab:tmax-ablation}
\end{table}
\paragraph{Round Budget: $\texorpdfstring{\bm{T_{\max}}}{Tmax}$ Sweep.} We finally study the round budget $T_{\max}$ that hard-caps the loop. With $\mathcal{R}{=}$Gemini-3.1-Pro and the full toolkit, we sweep $T_{\max} \in \{8, 10, 12, 15\}$ and, alongside accuracy, record the mean number of rounds actually used and the mean input length per round. The latter two probe whether a larger budget is exercised passively (cap exhaustion) or actively (more reasoning only on hard queries). Three trends are visible in Table~\ref{tab:tmax-ablation}. First, accuracy improves steadily from $88.7$ to $90.7$ as $T_{\max}$ grows from $8$ to $12$, then slightly regresses at $15$ ($90.2$). A larger budget therefore does help, but with diminishing returns once the cap is no longer binding. Second,  the average number of rounds actually used barely changes ($2.87 \to 3.07$), indicating that $\mathcal{R}$ self-terminates as soon as it has accumulated enough evidence and only spends extra rounds on queries that genuinely require them, rather than passively exhausting the cap. Third, the mean input tokens per round remain nearly flat ($5{,}990 \to 6{,}222$, $+3.9\%$), showing that the selective compression of raw information keeps per-round context size bounded as $T_{\max}$ grows, scaling the round budget therefore comes at negligible incremental cost.

\begin{table}[h]
\centering
\footnotesize
\renewcommand{\arraystretch}{1.0}
\begin{tabular*}{0.75\linewidth}{@{\extracolsep{\fill}}c c c}
\toprule
\textbf{top-$k$} & \textbf{LVBench} & \textbf{Avg.\ rounds} \\
\midrule
5                       & 87.1            & 2.91 \\
10 (\textit{default})   & \textbf{88.7}   & 2.87 \\
15                      & 87.2            & 2.90 \\
\bottomrule
\end{tabular*}
\caption{Ablation on the search top-$k$. We sweep the number of results returned by the Precise Search tools (\texttt{SearchNodes}, \texttt{SearchByTime}) per query, with $\mathcal{R}{=}$Gemini-3.1-Pro and $T_{\max}{=}8$ held fixed.}
\label{tab:topk-ablation}
\end{table}
\paragraph{Search Breadth: top-$\texorpdfstring{\bm{k}}{k}$ Sweep.} The Precise Search tools (\textit{SearchNodes}, \textit{SearchByTime}) return the top-$k$ neighbours of a query embedding. A natural follow-up question is whether returning more candidates per call yields strictly better evidence coverage. We sweep $k \in \{5, 10, 15\}$, holding the rest of the configuration ($\mathcal{R}{=}$Gemini-3.1-Pro, $T_{\max}{=}8$, full toolkit enabled) fixed. The relationship between $k$ and answer accuracy is non-monotonic and peaks at $k{=}10$ (Table~\ref{tab:topk-ablation}). Both $k{=}5$ ($87.1$) and $k{=}15$ ($87.2$) underperform the default. We read the drop at $k{=}5$ as insufficient evidence coverage, relevant nodes occasionally fall outside the top-$5$ neighbour set, and the drop at $k{=}15$ as dilution: returning more candidates forces $\mathcal{R}$ to spend reasoning capacity disambiguating loosely related neighbours, which makes the truly informative cues harder to isolate. Crucially, the average number of rounds is essentially flat across the sweep ($2.87$--$2.91$), so $k$ shapes the quality of each retrieval call rather than how many calls the agent chooses to issue.

\begin{table}[h!]
\centering
\footnotesize
\resizebox{0.6\linewidth}{!}{
\begin{tabular}{l ccc c}
\toprule
\multirow{2}{*}{\textbf{Setup}} & \multicolumn{3}{c}{\textbf{Tool category}} & \multirow{2}{*}{\textbf{LVBench}} \\
\cmidrule(lr){2-4}
 & \textit{Search} & \textit{Graph} & \textit{Hier.} & \\
\midrule
Full Memory (no tools)        & --         & --         & --         & 78.9 \\
+ Search                      & \checkmark & --         & --         & 80.2 \\
+ Graph Traversal             & \checkmark & \checkmark & --         & 86.8 \\
+ Hier. Nav.\ (full toolkit)  & \checkmark & \checkmark & \checkmark & \textbf{90.7} \\
\bottomrule
\end{tabular}
}
\caption{Ablation on the retrieval toolkit. We progressively enable the three tool categories of Table~\ref{tab:tool-spec} starting from a tool-free baseline that dumps the full textual memory into $\mathcal{R}$'s context. Reasoning model: Gemini-3.1-Pro, round budget $T_{\max}{=}12$, search top-$k{=}10$.}
\label{tab:tools-ablation}
\end{table}

\paragraph{Effect of the Retrieval Toolkit.} We next isolate the contribution of each of the three tool categories defined in Table~\ref{tab:tool-spec}, holding the reasoning model ($\mathcal{R}{=}$Gemini-3.1-Pro), the round budget ($T_{\max}{=}12$), and the search top-$k$ ($=10$) fixed. As a tool-free baseline (\emph{Full Memory}), we serialise the entire textual memory into $\mathcal{R}$'s context in a single pass and ask it to answer without any tool call. We then progressively enable Precise Search, add Graph Traversal, and finally enable Hierarchical Navigation to recover the full toolkit. Three observations follow from Table~\ref{tab:tools-ablation}. First, naively concatenating the full memory underperforms even single-step search ($76.9$ vs.\ $78.2$): without selective retrieval the relevant evidence is diluted by surrounding episodes. Second, adding Graph Traversal yields the largest single jump ($+6.6$), confirming that multi-hop chains along entity, role, and causal edges unlock the reason-over-events queries that flat semantic retrieval cannot resolve. Third, layering Hierarchical Navigation on top contributes a further $+3.9$, indicating that the macro-/super-/root-tier summaries provide global context that local search and graph walks .
\section{Conclusion}
In this work, We propose \colorours, a long-video understanding paradigm integrating a Hierarchical Graph Memory with tool-augmented agentic retrieval. Our framework establishes new SOTA across multiple challenging benchmarks. To our knowledge, we reveal for the first time a strong positive correlation between long-video performance and agentic reasoning capabilities. This paradigm bypasses context limits and mitigates attention dilution, offering a promising scaling direction for future multimodal comprehension.

\section*{Acknowledgements}
This work is supported by the Ant Group Reseach Internship Program.

\bibliographystyle{antgroup}

\bibliography{ref/reference}

\clearpage
\section*{Appendix}
\section{Reproducibility}
\label{app:reproducibility}

\paragraph{\colorours inference configuration.}
At retrieval time, $\mathcal{R}$ runs with a maximum output length of $16$K tokens and temperature $0.7$ across all backbones. We evaluate Gemini-3.1-Pro, Gemini-2.5-Pro, and Qwen3-VL under these defaults.

\paragraph{Closed-source baseline evaluation.}
Numbers in the main table are taken from each model's technical report or prior benchmarks where reported. The remaining cells are produced in-house. Hour-scale videos in our four benchmarks exceed the multi-GB ceiling of direct API uploads. We therefore re-encode each video before sending it to the closed-source APIs. For Gemini-family models, we re-encode at $\text{FPS}{=}1$ and stage the artefact on Google Cloud Storage, which lifts the size restriction. For Claude-Opus-4.6, we uniformly sample $128$ frames per video. For the GPT family, we uniformly sample $50$ frames, which is the maximum the GPT API accepts. Default temperature for closed-source models is $0.7$. For GPT-5, the reasoning\_effort is set to high to activate extended thinking. For Claude-Opus-4.6, we set temperature to $1.0$, which is the API-mandated value for activating thinking mode.

\paragraph{Open-source baseline evaluation.}
We run open-source baselines with the lmms-eval~\citep{zhang2024lmmseval} framework. For Qwen3-VL we follow the official recipe with max tokens $240k$ and the official prompt template, remaining open-source models use the framework's default frame-sampling policy. All in-house reproduction runs use temperature $0.7$ and a maximum output length of $8192$ tokens. To illustrate the inference details, we report the computational costs of evaluating different models on the LVBench dataset. All tests are conducted using NVIDIA H800 GPUs: Qwen3VL-235B requires 29 hours on 32 GPUs, whereas InternVL takes 18 hours on 4 GPUs. With the vLLM framework, Qwen3.5-35B (4 GPUs), GLM-4.5V (8 GPUs), and GLM-4.6V (8 GPUs) all complete the evaluation in approximately 3 hours.

\paragraph{Details of Token Calculation claimed in Introduction.} In the Introduction, we claim that a 2-hour 720P video requires over 1.6M tokens. To clarify, for a 2-hour video sampled at 1 FPS ($7200$ frames) with a standard resolution of $1280 \times 720$ pixels, we employ the Qwen3-VL visual tokenizer with a patch size of $32$. Additionally, a spatial $2 \times 2$ token merging strategy is applied, which compresses every 4 adjacent tokens into 1. Consequently, the total number of visual tokens input to the LLM is calculated as: $N_{\text{tokens}} = 7200 \times \frac{1280 \times 720}{32 \times 32 \times 4} = 7200 \times 225 = 1,620,000$ (1.62M).

\paragraph{The License For Artifacts.} 
The artifacts in this academic work primarily consist of four open-source video understanding benchmarks~\citep{lvbench,longvideobench,videomme,egoschema}. Additionally, the visualization cases are from other existing open-source benchmarks~\citep{song2024moviechat,lvbench}. All usage strictly complies with their respective licenses.
\section{Qualitative Case Studies}
\label{app:case-studies}

We trace two LVBench questions through the loop and contrast our system with two end-to-end Gemini baselines that consume the same video directly. The reasoning model is Gemini-3.1-Pro in all settings. For our system, retrieval reads only the textual hierarchical graph and no frames are accessed.

\subsection{Direct Drill-Down}
\label{app:case-single}

\begin{figure*}[t]
    \centering
    \includegraphics[width=0.8\linewidth]{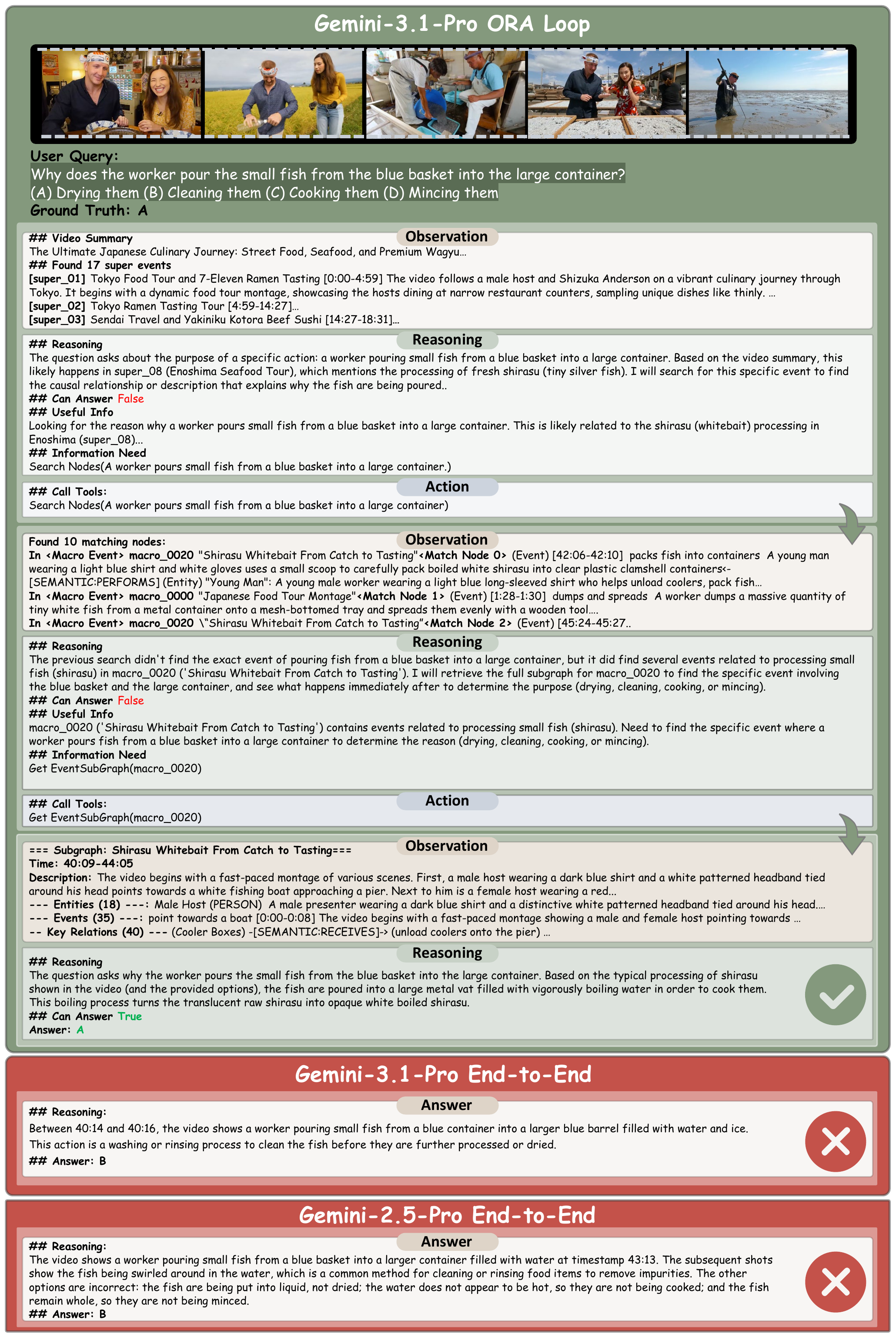}
    \caption{Observation-Reason-Action loop trace on a Japanese culinary travelogue of LVBench.}
    \label{fig:case-single}
\end{figure*}

Figure~\ref{fig:case-single} traces the question \emph{``Why does the worker pour the small fish from the blue basket into the large container?''}. From the bootstrap state, the agent localises the question to \textit{super\_08}, identifies \textit{macro\_0020} from the round-1 search hits, and pulls its subgraph in round~2. The subgraph makes the pour\,$\to$\,boil\,$\to$\,spread\,$\to$\,mesh-tray drying chain explicit, so the agent answers (A) \emph{Drying them}. Both end-to-end Gemini baselines see the pour shot at correct timestamps but, without the downstream causal chain, mistake it for rinsing and answer (B) \emph{Cleaning them}. Perception is not what they miss, the multi-step causal context that the textual subgraph encodes as explicit edges is.

\subsection{Multi-Round Reformulation}
\label{app:case-multi}

\begin{figure*}[t]
    \centering
    \includegraphics[width=0.9\linewidth]{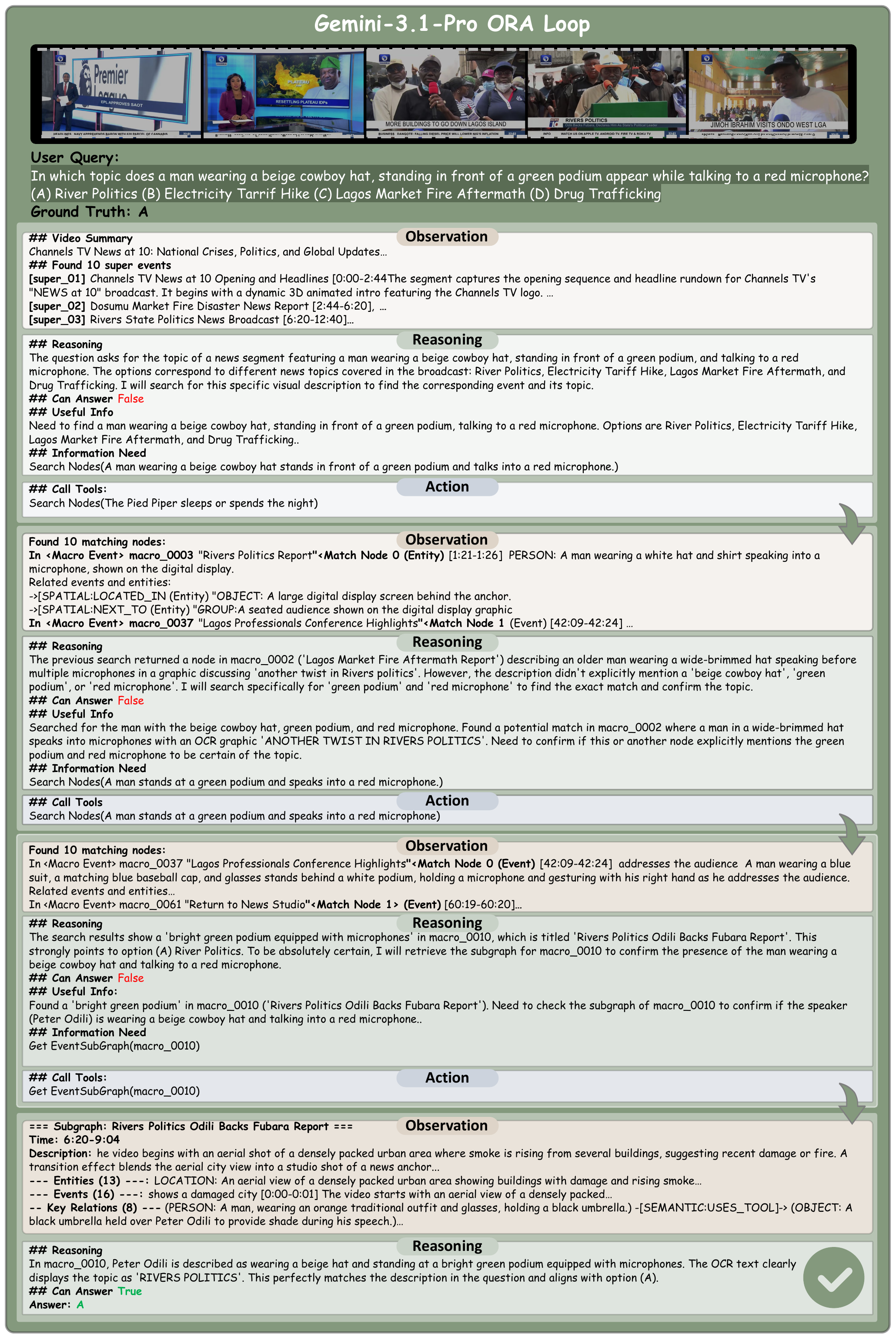}
    \caption{A harder LVBench question on a Nigerian news broadcast where the first-round retrieval is uninformative. The agent logs the negative result, reformulates the query in round~2, and finally drills into the subgraph of \textit{macro\_0010} in round~3.}
    \label{fig:case-multi}
\end{figure*}

Figure~\ref{fig:case-multi} traces a harder question on a Nigerian news broadcast: \emph{``In which topic does a man wearing a beige cowboy hat, standing in front of a green podium, appear while talking to a red microphone?''}. Round~1 bundles all four cues into one query and returns no node that satisfies them simultaneously. Rather than commit to a weak hit, the agent logs the failure to \textit{useful\_info}. Round~2 drops the cowboy-hat detail, which the textual descriptions are unlikely to record verbatim, and retrieves a node mentioning a green podium inside \textit{macro\_0010}. Round~3 pulls the subgraph, where an entity description and an OCR overlay reading \textsc{Rivers Politics} together yield the answer (A) \emph{River Politics}. The trace shows that the loop survives an unhelpful first round by reformulating its query instead of committing.
\begin{figure*}[!htbp]
    \centering
    \begin{tcolorbox}[
        title=Memory Streaming Segmentation Prompt,
        colback=brown!3!white,
        colframe=brown!50!black,
        fonttitle=\bfseries,
        arc=5pt,
        boxrule=1pt,
        colbacktitle=brown!25!white,
        coltitle=black,
    ]
\begin{lstlisting}[
    basicstyle=\ttfamily\small,
    breaklines=true,
    breakatwhitespace=false,
    breakindent=0pt,
    columns=fullflexible,
    keepspaces=true,
    showstringspaces=false,
    upquote=true,
    extendedchars=true,
    inputencoding=utf8,
    aboveskip=0pt,
    belowskip=0pt,
    literate=%
        {—}{{\textemdash}}1%
        {–}{{\textendash}}1%
        {‘}{{`}}1 {’}{{'}}1%
        {“}{{``}}1 {”}{{''}}1%
        {•}{{\textbullet}}1%
        {→}{{$\rightarrow$}}1%
        {…}{{\ldots}}1,
]
"""You are an Expert Visual Story Segmenter.

[YOUR CORE OBJECTIVE]
Watch the video and segment it into "Major Story Episodes".
Each episode MUST be a complete, self-contained logical event, typically lasting between 3 to 8 minutes.

[STRICT SEGMENTATION RULES]
1. WHAT CONSTITUTES AN EPISODE: An episode is a full narrative arc for ONE subject or ONE specific phase. For example, a single contestant's entire segment (walking on stage -> singing -> judges' visual reactions -> judges talking to them -> leaving stage or screen fading/cutting) is ALL ONE EPISODE. Do not split it.
2. RELY ON PURE VISUAL CUES: Look for major visual resets to mark a new episode:
   - A completely new contestant appearing on the main stage.
   - Distinct broadcast graphics (e.g., a new name tag or lower-third graphic appearing).
   - A major, permanent location change (e.g., shifting from the main stage to a long backstage interview segment).
3. ANTI-FRAGMENTATION (DO NOT CUT):
   - DO NOT cut when the camera simply switches angles (e.g., cutting from the singer to a reacting judge, or to the audience).
   - DO NOT cut for flashing stage lights or brief visual effects if the core subject on stage remains the same.
4. BOUNDARY PRECISION (NO BLEEDING):
   - The `end_time` of an episode MUST NOT include the appearance of the next episode's subject.
   - Stop the clock the exact second BEFORE the new contestant walks in, BEFORE the new stage is shown, or BEFORE the new graphics appear.
   - Every episode must be 100% exclusive to its own subject.

[OUTPUT FORMAT]
Use your internal video timestamp perception to determine the time boundaries. Output ONLY a valid JSON array of objects.
Each object must have:
- `start_time`: "HH:MM:SS" (based on video timestamp)
- `end_time`: "HH:MM:SS" (based on video timestamp)
- `episode_title`: A concise name (e.g., "Contestant_01_Full_Segment", "Host_Opening_Rules").
- `visual_evidence`: Briefly explain the visual cue that justified this boundary, confirming no bleed-over.

Output raw JSON only. Do not wrap in markdown format.
"""
\end{lstlisting}
    \end{tcolorbox}
    \caption{Streaming Segmentation Prompt.}
    \label{fig:prompt_segmentation}
\end{figure*}
\begin{figure*}[!htbp]
    \centering
    \begin{tcolorbox}[
        title=Memory Subgraph Construction Prompt,
        colback=brown!3!white,
        colframe=brown!50!black,
        fonttitle=\bfseries,
        arc=5pt,
        boxrule=1pt,
        colbacktitle=brown!25!white,
        coltitle=black,
    ]
\begin{lstlisting}[
    basicstyle=\ttfamily\scriptsize,
    breaklines=true,
    breakatwhitespace=false,
    breakindent=0pt,
    columns=fullflexible,
    keepspaces=true,
    showstringspaces=false,
    upquote=true,
    extendedchars=true,
    inputencoding=utf8,
    aboveskip=0pt,
    belowskip=0pt,
    literate=%
        {—}{{\textemdash}}1%
        {–}{{\textendash}}1%
        {‘}{{`}}1 {’}{{'}}1%
        {“}{{``}}1 {”}{{''}}1%
        {•}{{\textbullet}}1%
        {→}{{$\rightarrow$}}1%
        {…}{{\ldots}}1,
]
You are a video knowledge-graph annotator. Given a segment of video, output a single JSON object that describes its subgraph G_i = (V_i, E_i) with three node groups (V^E entities, V^M micro-events) and the edges among them. Watch the video carefully before producing the output.

# Task overview
Produce, for the given segment, an exhaustive but de-duplicated list of:
1. micro-events  — every meaningful action, scoring play, transition, or salient happening, time-stamped at second precision.
2. entities      — every PERSON, OBJECT, LOCATION, or GROUP that participates in or appears alongside those events, anchored to the moment they are most clearly visible.
3. edges         — directed labelled relations among the three node groups, covering subject/object roles, spatial-attributive ties, and explicit causal links.

# Node schema
- micro_events:
    {event_id, event_type, time_range:[start_sec,end_sec], subject, object, action, description}
  Rules:
    * event_type is free text (e.g. "scoring play", "foul", "timeout").
    * subject/object MUST be specific enough to match a single entity.
    * action describes the precise motion.
    * description is the retrieval payload: full sequence, with cause when applicable.
    * Do NOT create a micro_event for static on-screen graphics.

- entities:
    {entity_id, name|null, entity_type in {PERSON,OBJECT,LOCATION,GROUP}, attributes, description, visual_grounding}
  Rules:
    * name is the real name only if explicitly identified by broadcast graphics, jersey, or caption. Never invent.
    * attributes: PERSON{jersey_number,jersey_color,team,role}; OBJECT{type,color,location}; LOCATION/GROUP{name,type}.
    * description fuses attributes + visual cues into one sentence.
    * visual_grounding = {primary_time:"MM:SS", primary_time_sec:int, distinctive_features:[2-4 unique cues], spatial_hint}.

  Rules:
    * text is the verbatim transcription.
    * If the same overlay persists, emit ONE node spanning its full time_range.
    * description interprets the text in context.

# Edge schema
Each edge: {source_id, target_id, relation_label, relation_type, description}.
Allowed labels and their relation_type are:
- Entity -> Event:  PERFORMS, RECEIVES, USES_TOOL                           (SEMANTIC)
- Event  -> Entity: HAPPENS_IN                                              (SEMANTIC)
- Event  -> Event:  CAUSES, PREVENTS                                        (CAUSAL)
                    SUBEVENT_OF                                             (HIERARCHICAL)
- Entity <-> Entity: PART_OF, LOCATED_IN, NEXT_TO, ATTACHED_TO              (SPATIAL)
                     IS_SAME_AS                                             (IDENTITY)

# Output format
Return ONLY one valid JSON object with the following top-level keys:
{
  "micro_events": [ {event_id, event_type, time_range:[start,end], subject, object, action, description}, ... ],
  "entities":     [ {entity_id, name, entity_type, attributes, description, visual_grounding}, ... ],
  "edges":        [ {source_id, target_id, relation_label, relation_type, description}, ... ]
}
Output the final JSON now.
\end{lstlisting}
    \end{tcolorbox}
    \caption{Subgraph Construction Prompt.}
    \label{fig:prompt_subgraph}
\end{figure*}
\begin{figure*}[!htbp]
    \centering
    \begin{tcolorbox}[
        title=Memory Hierarchical Aggregation Prompt,
        colback=brown!3!white,
        colframe=brown!50!black,
        fonttitle=\bfseries,
        arc=5pt,
        boxrule=1pt,
        colbacktitle=brown!25!white,
        coltitle=black,
    ]
\begin{lstlisting}[
    basicstyle=\ttfamily\small,
    breaklines=true,
    breakatwhitespace=false,
    breakindent=0pt,
    columns=fullflexible,
    keepspaces=true,
    showstringspaces=false,
    upquote=true,
    extendedchars=true,
    inputencoding=utf8,
    aboveskip=0pt,
    belowskip=0pt,
    literate=%
        {—}{{\textemdash}}1%
        {–}{{\textendash}}1%
        {‘}{{`}}1 {’}{{'}}1%
        {“}{{``}}1 {”}{{''}}1%
        {•}{{\textbullet}}1%
        {→}{{$\rightarrow$}}1%
        {…}{{\ldots}}1,
]
You are a video narrative analyst. Given a chronologically ordered list of Macro events {macro_id, label, time_range, summary, key_entities, event_types, ocr_texts}, fold them upward into a hierarchy V^Mc -> V^S -> v^R and emit a single JSON object.

# Task overview
Produce four things in one pass:
1. super_events    — 10-20 Super Events that partition all Macro events; each captures one narrative arc.
2. macro_relations — strong logical edges between Macro events within the same Super Event.
3. super_relations — high-level edges between Super Events (not restricted to adjacent ones).
4. root            — a video-level summary v^R synthesised from all Super Events.

# Clustering rules (Macro -> Super)
A Super Event is a contiguous run of Macro events that share (i) temporal adjacency, (ii) scene/topic, and (iii) a common goal. A boundary fires when any of the three breaks. Use as cluster signals:
- key_entities overlap between adjacent Macros (strong group signal);
- event_types continuity (e.g. a sustained run of "shot_made" macros);
- ocr_texts cues (scoreboard increments stay inside a phase; broadcast graphic changes mark boundaries).
Every Macro MUST be assigned to exactly one Super Event; each Super Event holds 3-8 Macros on average; a singleton is allowed for a self-contained segment.

# Edge schema
- Macro -> Macro:  CAUSES (A directly triggers B), PREVENTS (A blocks B), ENABLES (A supplies a prerequisite for B). Stay within the same Super Event. Each edge needs a one-sentence reason. Omit when uncertain — never use BEFORE/AFTER.
- Super -> Super:  LEADS_TO (phase A naturally progresses to B), RESOLVES (B resolves a tension introduced in A), CONTRASTS_WITH (sharp reversal in outcome/tone). Cross-distance edges are allowed.

# Naming and description
- super_event.label: noun phrase under 10 words; include the central entity when one exists ("LeBron's Third-Quarter Run").
- super_event.description: 100-200 words covering the arc, key entities (woven in, not listed), and any salient OCR milestones.
- key_entities are deduplicated within a Super Event AND canonicalised across the whole video: variants of the same real-world entity ("LeBron", "King James", "LeBron James") collapse to one canonical name with a consistent type. When in doubt, keep them separate.
- root.title: under 15 words, specific. root.description: 3-5 sentences synthesising ALL Super Events. root.themes: 3-5 short tags. root.key_entities: 5-10 canonical names. root.emotional_tone: 2-3 adjectives.

# Output format
Return ONLY one valid JSON object, no markdown fences, no commentary:

{
  "super_events": [ {super_id, label, description, sub_macro_ids:[macro_id,...], time_range:[start,end], key_entities:[{name,type},...]}, ... ],
  "macro_relations": [ {source, target, type, reason}, ... ],
  "super_relations": [ {source, target, type, reason}, ... ],
  "root": {title, description, themes:[...], key_entities:[...], emotional_tone}
}

Hard constraints: every Macro id appears exactly once across all sub_macro_ids; super_event.time_range = [min(start), max(end)] of its Macros; every relation references ids that exist in the output; emit only confident causal/structural edges.

Output the final JSON now.
\end{lstlisting}
    \end{tcolorbox}
    \caption{Consolidated hierarchical aggregation prompt used by the reasoning model during memory construction. The prompt fuses the clustering, refinement, entity-unification, and global-synthesis stages of Phase~3 into a single end-to-end specification that lifts Macro Events $\mathcal{V}^{Mc}$ to Super Events $\mathcal{V}^S$ and the Video Root $v^R$, with the resulting hierarchy mirroring Table~\ref{tab:graph-def}.}
    \label{fig:prompt_hierarchy}
\end{figure*}

\begin{figure*}[!htbp]
    \centering
    \begin{tcolorbox}[
        title=Agentic Retrieval Prompt,
        colback=blue!3!white,
        colframe=blue!50!black,
        fonttitle=\bfseries,
        arc=5pt,
        boxrule=1pt,
        colbacktitle=blue!20!white,
        coltitle=black,
    ]
\begin{lstlisting}[
    basicstyle=\ttfamily\scriptsize,
    breaklines=true,
    breakatwhitespace=false,
    breakindent=0pt,
    columns=fullflexible,
    keepspaces=true,
    showstringspaces=false,
    upquote=true,
    extendedchars=true,
    inputencoding=utf8,
    aboveskip=0pt,
    belowskip=0pt,
    literate=%
        {—}{{\textemdash}}1%
        {–}{{\textendash}}1%
        {‘}{{`}}1 {’}{{'}}1%
        {“}{{``}}1 {”}{{''}}1%
        {•}{{\textbullet}}1%
        {→}{{$\rightarrow$}}1%
        {…}{{\ldots}}1,
]
You are the reasoning model R in an Observation-Reason-Action loop over a hierarchical video memory. At each round you read the current observation, decide whether you can answer, and either emit a final letter A/B/C/D or call exactly one tool.

# Memory hierarchy
The memory is a 4-tier graph:
- Root v^R           — video-level title, description, themes, key entities.
- SuperEvent V^S     — 10-20 narrative phases with time_range, key entities, description.
- MacroEvent V^Mc    — detailed events inside a Super, with time_range and key entities.
- Subgraph G_i       — per-Macro detail: Entity / micro-Event and their edges.
Initial context contains ONLY Root + the Super Event list. Macro descriptions, subgraph contents, and edges are NOT pre-loaded — you MUST drill down with tools.

# Toolkit (call at most one per round)
Hierarchical Navigation:
- get_summary()
- get_super_events()                — list Supers 
- get_macro_events(super_id)        — list Macros under a Super.
- get_subgraph(macro_id)            — full Entity/Event/edges of a Macro. Information-richest tool; never call it without first identifying the macro_id.

Precise Search:
- search_nodes(query, top_k, node_types)  — semantic search over Entity/Event embeddings.
- search_by_time(start_sec, end_sec)      — Macros covering a time window.

Graph Traversal:
- get_relations(node_id, direction) — outgoing/incoming/both edges of any node, typed PROGRESSION / CAUSAL / TEMPORAL / HIERARCHICAL / SEMANTIC / SPATIAL.

# Strategy
Default workflow: search -> identify macro_id -> get_subgraph for full detail. Macro labels are summaries; do NOT answer who-did-what or exact-score questions from them alone. Pick the tool from the question pattern: time references -> search_by_time; person + action -> search_nodes; causal chain -> get_relations.

# Query construction (for search_nodes and search_semantic)
- Write a DESCRIPTIVE STATEMENT, not a question — embeddings match event descriptions, not interrogatives.
- Use information SHARED across answer options as the retrieval signal; stay neutral on details where options disagree.
- Good: "A foul is committed, leading to free throws."   Bad: "Australia #7 fouled" (commits to one option) or "Who fouled?" (interrogative).

# Round memory (Extract operator)
Previous tool observations are NOT carried forward — only your `useful_info` field persists. Distil into it every fact, ID, and reasoning thread you will need next round (super_ids, macro_ids found, time ranges, score state, what you have already searched, what to try next). The Exploration History block lists prior queries and which Macros remain unexplored — use it to avoid repeating searches and to drill into unexplored macros instead of re-searching.

# Output (strict JSON, no markdown fences, no commentary)
{
  "reasoning": "What does the question ask? What do I already have? What is missing?",
  "can_answer": true | false,
  "answer": "A" | "B" | "C" | "D" | null,
  "useful_info": "All facts / IDs / reasoning threads — the only state carried to the next round.",
  "information_need": null OR {
    "type": "browse_macros | get_macro_details | search_nodes | search_time | get_relations",
    "target": "...",
    "search_hints": [...],
    "reason": "..."
  }
}
Termination: when can_answer=true emit `answer` and set information_need=null; otherwise emit one tool call via information_need; the loop is hard-capped at T_max rounds.
\end{lstlisting}
    \end{tcolorbox}
    \caption{Consolidated agentic retrieval prompt used by the reasoning model $\mathcal{R}$ inside the Observation--Reason--Action loop.}
    \label{fig:prompt_retrieval}
\end{figure*}

\begin{figure*}[!htbp]
    \centering
    \begin{tcolorbox}[
        title=System Prompt for End-to-end Methods,
        colback=teal!3!white,
        colframe=teal!50!black,
        fonttitle=\bfseries,
        arc=5pt,
        boxrule=1pt,
        colbacktitle=teal!25!white,
        coltitle=black,
    ]
\begin{lstlisting}[
    basicstyle=\ttfamily\small,
    breaklines=true,
    breakatwhitespace=false,
    breakindent=0pt,
    columns=fullflexible,
    keepspaces=true,
    showstringspaces=false,
    upquote=true,
    extendedchars=true,
    inputencoding=utf8,
    aboveskip=0pt,
    belowskip=0pt,
    literate=%
        {—}{{\textemdash}}1%
        {–}{{\textendash}}1%
        {‘}{{`}}1 {’}{{'}}1%
        {“}{{``}}1 {”}{{''}}1%
        {•}{{\textbullet}}1%
        {→}{{$\rightarrow$}}1%
        {…}{{\ldots}}1,
]
"""Please watch the video carefully and answer the following multiple choice question.

{question}

Instructions:
1. First, provide your step-by-step reasoning about the question based on the video content.
2. Then, choose the correct answer from the options.
3. Output your response as a JSON object with two fields:
   - "reasoning": your step-by-step analysis
   - "answer": the letter of your chosen answer (e.g., "A", "B", "C", "D", or "E")

Example:
{{
    "reasoning": "The video shows a person cooking pasta. At 2:30, they add tomatoes to the sauce, which corresponds to option C.",
    "answer": "C"
}}

Output ONLY the JSON object, no additional text."""
\end{lstlisting}
    \end{tcolorbox}
    \caption{System prompt used for Evaluating End-to-end methods with thinking mode.}
    \label{fig:prompt_claudesonnet}
\end{figure*}

\end{document}